\documentclass[letterpaper]{article} 
\usepackage[]{aaai2026} 
\usepackage{times} 
\usepackage{helvet} 
\usepackage{courier} 
\usepackage[hyphens]{url} 
\usepackage{graphicx} 
\usepackage{natbib} 
\usepackage{caption} 
\usepackage{algorithm}
\usepackage{algorithmic}
\usepackage{times}
\usepackage{latexsym}
\usepackage{amsmath}
\usepackage{booktabs}
\usepackage{enumitem}
\usepackage{verbatim}
\usepackage{placeins}
\usepackage{tabularx}
\usepackage{newfloat}
\usepackage{listings}
\DeclareCaptionStyle{ruled}{labelfont=normalfont,labelsep=colon,strut=off} 
\floatstyle{ruled}
\newfloat{listing}{tb}{lst}{}
\floatname{listing}{Listing}
 \nocopyright 

\title{Dissecting Clinical Reasoning in Language Models:\\ A Comparative Study of Prompts and Model Adaptation Strategies}
\author{
 \textbf{Maël Jullien\textsuperscript{1}\textsuperscript{,3}, Marco Valentino\textsuperscript{4}, Leonardo Ranaldi\textsuperscript{5},   Andr\'e Freitas\textsuperscript{1}\textsuperscript{,2}\textsuperscript{,3}}\\ 
}
\affiliations{
 $^{1}$Department of Computer Science, University of Manchester, UK, \\ 
$^{2}$ National Biomarker Centre, CRUK-MI, University of Manchester, UK\\
$^{3}$Idiap Research Institute, Switzerland, $^{4}$ School of Computer Science, University of Sheffield, UK, \\$^{5}$ School of Informatics, University of Edinburgh, UK\\
$^{3}${\tt \{firstname.surname\}\tt@idiap.ch}}

\begin{document}
\maketitle
\begin{abstract}

Recent works on large language models (LLMs) have demonstrated the impact of prompting strategies and fine-tuning techniques on their reasoning capabilities. Yet, their effectiveness on \emph{clinical natural language inference} (NLI) remains underexplored. This study presents the first controlled evaluation of how prompt structure and efficient fine-tuning jointly shape model performance in clinical NLI.

We inspect four classes of prompting strategies to elicit reasoning in LLMs at different levels of abstraction, and evaluate their impact on a range of clinically motivated reasoning types. For each prompting strategy, we construct high-quality demonstrations using a frontier model to distil multi-step reasoning capabilities into smaller models ($\leq$4B parameters) via Low-Rank Adaptation (LoRA). Across different language models fine-tuned on the NLI4CT benchmark, we found that prompt type alone accounts for up to 44\% of the variance in macro-F\textsubscript{1}. Moreover, LoRA fine-tuning yields consistent gains of +8–12 F\textsubscript{1}, raises output alignment above 97\%, and narrows the performance gap to GPT-4o-mini to within 7.1\%. Additional experiments on reasoning generalisation reveal that LoRA improves performance in 75\% of the models on MedNLI and TREC Clinical Trials Track.

Overall, these findings demonstrate that (i) prompt structure is a primary driver of clinical reasoning performance, (ii) compact models equipped with strong prompts and LoRA can rival frontier-scale systems, and (iii) reasoning-type-aware evaluation is essential to uncover prompt-induced trade-offs. Our results highlight the promise of combining prompt design and lightweight adaptation for more efficient and trustworthy clinical NLP systems, providing insights on the strengths and limitations of widely adopted prompting and parameter-efficient techniques in highly specialised domains\footnote{All code, annotations, prompts, demonstrations, and checkpoints will be released upon publication.}.
\end{abstract}

\section{Introduction}
\begin{figure*}[!t]
 \centering
 \includegraphics[width=0.7\textwidth]{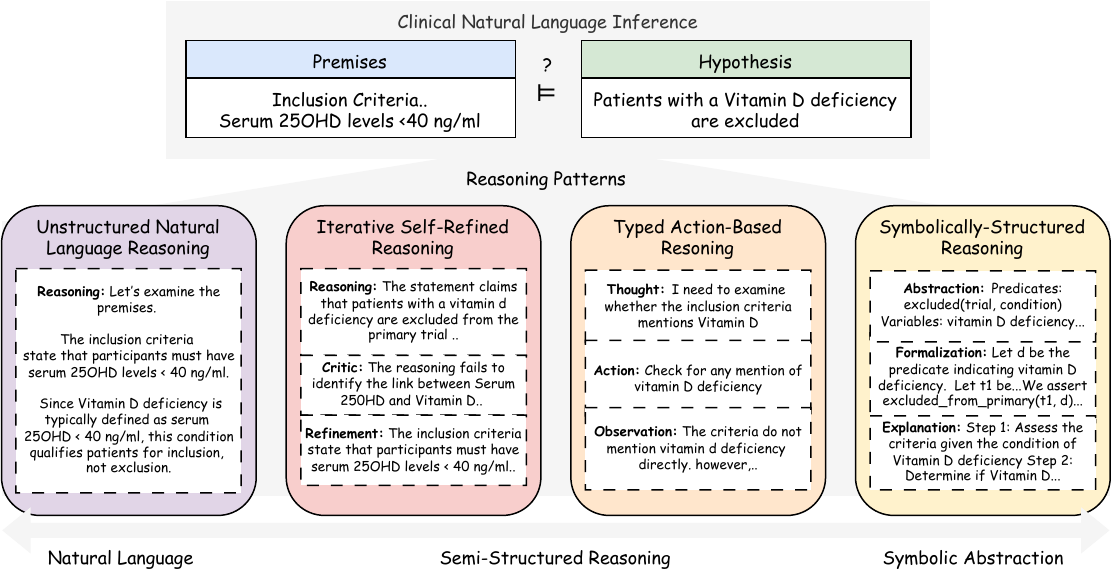}
\caption{Example reasoning trajectories for a single NLI4CT instance under four prompting strategies: NLR, ISRR, TAR, and SSR.
The prompts are shown in the context of our five-stage experimental framework:
\textbf{(1)} Reasoning types defined and annotated;
\textbf{(2)} Prompt categories defined and instantiated as structured scaffolds;
\textbf{(3)} Generation of high-precision demonstrations;
\textbf{(4)} Compact models adapted via LoRA on prompt-aligned data;
\textbf{(5)} Evaluation performed by reasoning type and benchmark.} 

 \label{fig:main}
\end{figure*}

LLMs now achieve state-of-the-art performance across a broad spectrum of reasoning tasks, including mathematics, commonsense, and science~\citep{brown2020language, bommasani2021opportunities, achiam2023gpt}. However, LLM performance is closely correlated to prompt design. A growing body of work has shown that different prompting strategies, such as chain-of-thought~\citep{wei2022chain}, self-ask~\citep{press2022measuring}, and ReAct~\citep{yao2023react}, can yield significant variation in reasoning behaviour~\citep{zhang2022automatic, wen2025thinkpatterns, mondorf2024beyond}.

Prompt engineering has been applied to clinical tasks such as medical question answering~\citep{nori2023can}, scientific claim verification~\citep{ma2023sci}, and natural language inference (e.g., NLI4CT)~\citep{jullien2023nli4ct}. However, prior works on NLI4CT have used heterogeneous model architectures, prompting strategies, and tuning protocols, precluding a systematic comparison. As a result, it remains unclear which inference strategy best supports the composite reasoning skills required in the clinical domain.

This paper presents the first systematic evaluation of prompting and parameter-efficient adaptation for clinical NLI under controlled conditions. By holding model scaling ($\geq$4B), and training configuration, our experimental setup enables direct comparison across five structurally distinct prompting paradigms (Figure~\ref{fig:main}). To support interpretable analysis, we introduce expert-labeled reasoning annotations for the NLI4CT test set, spanning six categories: Clinical, Lexical Equivalence, World Knowledge, Expectation-Driven Evidence, Quantitative Comparison, and Quantitative Derivation. 

Our experiments reveal the following key findings:

\paragraph{Prompt structure is a primary driver of clinical reasoning performance.} Prompt structure alone explains up to \textbf{44 \%} of the variance in F\textsubscript{1}. This reveals the importance of prompt design for clinical NLI, and, at the same time, the persisting sensitivity of LLMs to specific input instructions.

\paragraph{Compact models equipped with strong prompts and LoRA can rival frontier-scale systems.} 
LoRA supplies consistent gains of \textbf{+8–12 F\textsubscript{1}} and lifts answer validity above 97\%, allowing a 3.8B model to trail GPT-4o-mini by 7.1\%. This shows that parameter-efficient adaptation techniques using high-quality demonstrations are a viable solution to reduce the gap between frontier proprietary models and smaller open-source models in specialised domains.

\paragraph{Reasoning-type-aware evaluation is essential to uncover prompt-induced trade-offs.} Prompt strategy explains 30\% to 44\% of variance across reasoning types, when controlling for model architecture and LoRA. This demonstrates the importance of fine-grained reasoning evaluation when assessing inference strategy in the clinical domain.

\paragraph{LoRA fine-tuning can generalise to different clinical NLI tasks.} We found significant improvements on MedNLI and the TREC Clinical Trials Track when fine-tuning on NLI4CT alone,
where LoRA boosts F\textsubscript{1} in 75 \% of model–prompt settings. Moreover, we found that the gap between fine-tuned models and GPT-4 is particularly reduced on more complex NLI tasks (i.e., TREC). This demonstrates that some of the inference capabilities acquired via parameter-efficient fine-tuning are preserved across distribution and complexity variations of the clinical NLI tasks.


\section{Methodology}

\subsection{Methodology Overview}

This study follows a five-stage framework illustrated in Figure~\ref{fig:main}. \textbf{(1) Reasoning Type Definition.} A typology of six reasoning types essential for clinical inference is defined. Each instance in the NLI4CT test set is manually annotated with one or more reasoning labels. \textbf{(2) Prompt Strategy Design.} A taxonomy of four abstract prompt categories is defined, each representing a distinct structural mode of inference. One representative prompt is constructed per category. \textbf{(3) Demonstration Collection.} For each prompt strategy, validated demonstrations are collected by applying GPT-4o-mini to the NLI4CT training set and retaining only instances for which the model prediction matches the gold label. \textbf{(4) LoRA Adaptation.} Four target models in the 1.5--3.8B parameter range are fine-tuned across all demonstration sets independently, using LoRA. \textbf{(5) Evaluation and Analysis.} All adapted models are evaluated on the full reasoning-annotated NLI4CT test set, as well as two out-of-domain generalisation benchmarks: MedNLI and TREC.

\subsection{Reasoning Types in Clinical Trial NLI}
\label{sec:reasoning}
Six reasoning categories are defined for NLI in this setting: \textit{Clinical}, \textit{Lexical Equivalence}, \textit{Evidence}, \textit{World Knowledge}, \textit{Quantitative Comparison}, and \textit{Quantitative Derivation}. Each category isolates a distinct inferential mechanism, that can be applied simultaneously, or concurrently within a single inference. Table~\ref{tab:reasoning_with_steps} presents representative examples for each category, which are formally defined below.

\begin{table*}[h!]
\scriptsize
\begin{tabular}{p{2.5cm} p{3cm} p{4cm} p{6.5cm}}
\toprule
\textbf{Reasoning Type} & \textbf{Statement} & \textbf{Premise} & \textbf{Reasoning} \\
\midrule
\raggedright \textbf{Clinical Reasoning} & One emesis episode and ondansetron given. & Vomiting was controlled. & Controlled = \emph{no} vomiting and \emph{no} rescue meds. Rescue therapy disqualifies control. $\Rightarrow$ Contradiction \\
\midrule
\raggedright \textbf{Lexical Equivalence} & Participants must have failed platinum therapy. & Inclusion Criteria: 
Previous treatment with Cis/Gem that didn’t control disease &  (``Cis/Gem''  $\Rightarrow$  "platinum-based regimens" \& ``Failed therapy''  $\Rightarrow$ ``disease not controlled'') $\Rightarrow$ Entailment. \\
\midrule
\raggedright \textbf{Expectation-Driven Evidence Reasoning} & The primary trial has 2 separate cohorts & Intervention section: once-daily oral dose of empagliflozin 10mg & Expected cohort/phase names or distinct treatment paths; none found (closed-world assumption) $\Rightarrow$ Contradiction. \\
\midrule
\raggedright \textbf{World‐Knowledge Inference} & The trial excludes children. & Only participants 18 years or older may enroll. & Age $\geq$ 18 implies participants are adults, thus excluding children $\Rightarrow$ Entailment. \\
\midrule
\raggedright \textbf{Quantitative Comparison} & Her CrCl is below 30 mL/min. & Creatinine clearance calculated as 28 mL/min. & 28 $<$ 30 $\Rightarrow$ Entailment. \\
\midrule
\raggedright \textbf{Domain-Grounded Quantitative Derivation} & All C1 patients receive higher ALT doses than C2 patients. & Cohort 1: ALT-801 0.04 mg/kg Cohort 2: ALT-801 0.01 mg/kg & Min C1 weight 40kg: $0.04 \times 40 = 1.6$mg. Max C2 weight 150kg: $0.01 \times 150 = 1.5$mg. $1.6>1.5$ $\Rightarrow$ Entailment. \\
\bottomrule
\end{tabular}
\caption{Examples of inference steps for each reasoning type on NLI4CT examples.}
\label{tab:reasoning_with_steps}
\end{table*}

\begin{table*}[t]
\scriptsize
\begin{tabular}{@{}p{3cm}p{5.5cm}p{3.5cm}p{4.5cm}@{}}
\toprule
\textbf{Category} & \textbf{Structural Signature} & \textbf{Reasoning Trajectory} & \textbf{Example Prompts} \\
\midrule
\raggedright\textbf{Unstructured Natural Language Reasoning (NLR)} & 
Free-form explanation followed by a final answer, with no explicit phases or typed components. & \raggedright
Single contiguous segment: $r_1, r_2, \dots, r_k \rightarrow y$ & 
\textit{“Let’s think step by step”}, \textit{“Explain your reasoning”} \textit{“Describe your thought process."} \\
\midrule
\raggedright\textbf{Iterative Self-Refined Reasoning (ISRR)} & 
Multi-stage reasoning with internal self-evaluation. An initial response, a critique or verification, and a revised answer. & 
\begin{tabular}[t]{@{}l@{}}
Stage 1: $y'$ (draft) \\
Stage 2+: $c$ (critique) \\ Stage 3+: $y$ (revised)
\end{tabular} & 
\textit{“Verify your reasoning at every step”}, \textit{“Self-debate between supportive and critical perspectives.”} 
\\
\midrule
\raggedright\textbf{Typed Action-Based Reasoning (TAR)} & 
Natural language interleaved with discrete, typed actions (e.g., \texttt{SEARCH}, \texttt{LOOKUP}) that return observable results. & 
\begin{tabular}[t]{@{}l@{}}
$(\text{Thought}_t, \text{Action}_t, \text{Obs}_t)_{t=1}^{T}$ \\ $\rightarrow y$
\end{tabular}
 & 
\textit{“Use Search\_PUBMED(query) to look up...”}, \textit{“Use QUERY\_DB(sql) to retrieve ...”}\\
\midrule
\raggedright\textbf{Symbolically Structured Reasoning (SSR)} & 
An intermediate symbolic representation (e.g., predicate logic, code, or JSON object) before deriving an answer. & 
$X \xrightarrow{\text{Abstract}} S \xrightarrow{\text{Solve}} y$, where $S$ belongs to a formal language & 
\textit{“Translate this to predicate logic”}, \textit{“Write a python function to solve..”} \\
\bottomrule
\end{tabular}
\caption{Illustration of reasoning trajectories across four prompting categories (NLR, TAR, ISRR, SSR), including their structural signatures, control flow patterns, and representative prompt examples.}
\label{tab:prompt_structure}
\end{table*}

\paragraph{1. Clinical Trial Reasoning}
Application of clinical knowledge within the context of a clinical trial protocol. This reasoning interprets clinical concepts (labs, events, terminology) through the lens of operational definitions, regulatory frameworks, and trial procedures. As a result, common clinical terms and expressions often take on specialized meanings that diverge from general interpretations.

\paragraph{2. Lexical Equivalence}
Identifying semantically equivalent phrases, based on surface‐level lexical synonymy.

\paragraph{3. Expectation-Driven Evidence Reasoning}
The statement is decomposed into a set of evidence markers, a discrete set of expectations about structural form, functional role, or factual content (e.g., actions, entities, measurements, or reported outcomes) that should appear in the premise if the statement holds. The premise is then examined for content that aligns with the expected form and function of these markers, permitting partial evidence. Under a closed-world assumption, the absence of the expected evidence is treated as evidence of negation. 

\paragraph{4. World‐Knowledge Inference}
Applying general world knowledge, and common sense rules. 

\paragraph{5. Quantitative Comparison}
Direct numeric entailment by single step comparison of values and thresholds explicitly defined in the premise and statement. 

\paragraph{6. Domain-Grounded Quantitative Derivation}
Multi‐step arithmetic on variables that may be directly specified, contextually inferred, or estimated using domain knowledge.

\subsection{Prompt Categories as Structural Abstractions}
\label{sec:prompt-taxonomy}

We introduce a taxonomy of four abstract prompt categories, each corresponding to a distinct structural mode of reasoning. These categories span a continuum from fully natural language to semi-symbolic abstraction. This taxonomy frames prompts as inductive biases that shape the reasoning trajectory by imposing structural scaffolds on how information is processed, decomposed, and justified. Definitions are provided in Table~\ref{tab:prompt_structure} and illustrated in Figure \ref{fig:main}.

\subsection{Automatic Demonstration Generation}
Demonstrations are automatically generated by a frontier model applied to a source corpus. Predictions are compared to gold-standard labels, and only correctly classified instances are retained, yielding high-precision, label-faithful, and task-aligned demonstration sets.

\subsection{LoRA Fine-Tuning}
Parameter-efficient adaptation is performed using LoRA \cite{hu2022lora}, which injects trainable low-rank adapters into the attention and feed-forward layers while keeping all other parameters fixed. This enables consistent, efficient fine-tuning across models and prompt strategies, preserving core model representations and facilitating controlled evaluation of prompt sensitivity and generalisation. Fine-tuning is supervised using a fixed number of randomly sampled demonstrations per prompt type, selected to balance convergence stability and overfitting risk. A uniform hyperparameter configuration is applied across models for comparability.

\begin{figure*}[t]
 \centering
 \includegraphics[width=0.9\textwidth]{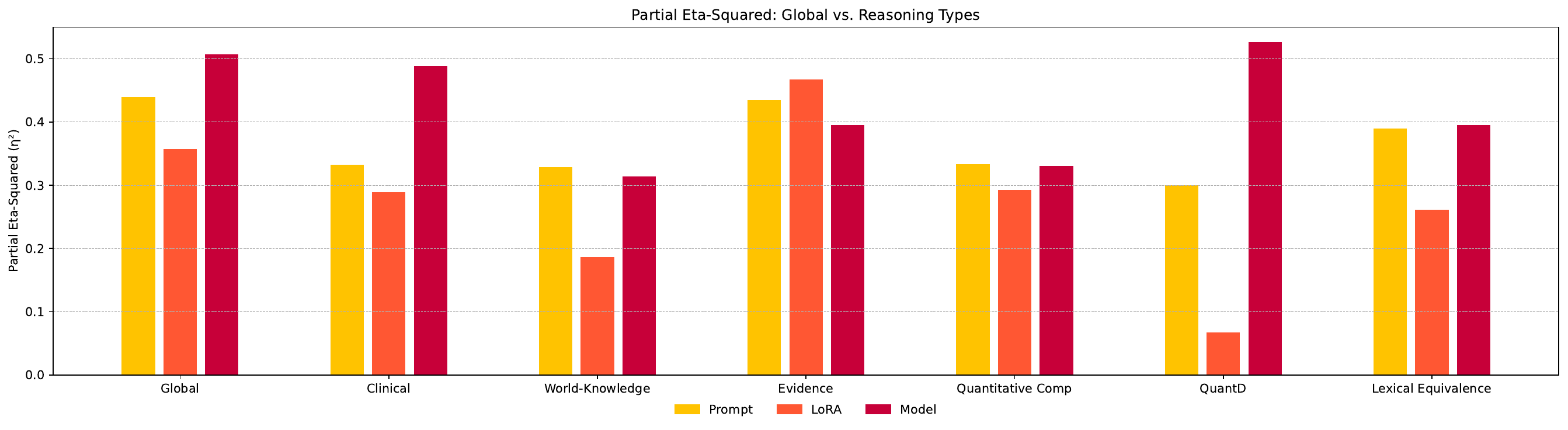}
 \caption{Partial $\eta^2$ values showing the proportion of variance in F$_1$ performance explained by Prompt, LoRA adaptation, and Model architecture, both globally and across reasoning types.}
\label{fig:eta}
\end{figure*}

\begin{figure}[t]
 \centering
 \includegraphics[width=\columnwidth]{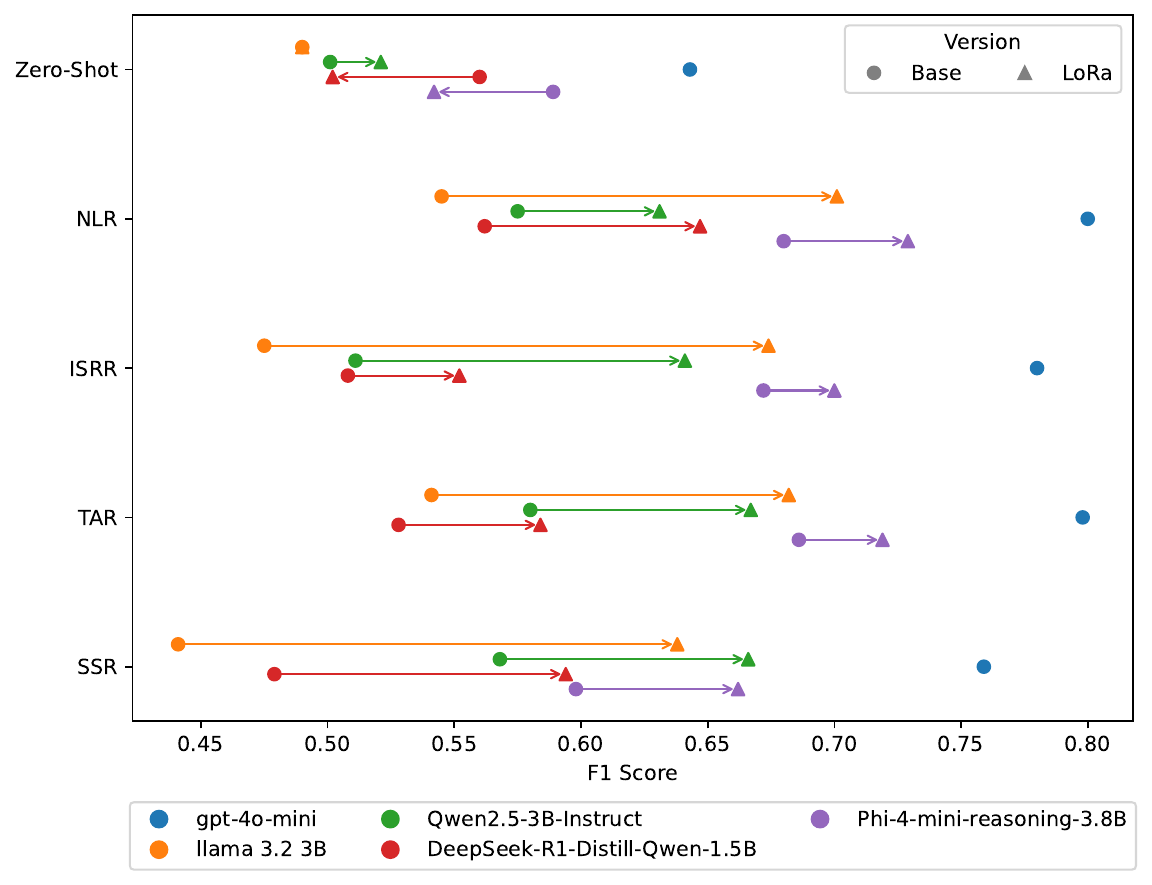}
 \caption{Macro F\textsubscript{1} Scores on the NLI4CT Test Set}
\label{fig:nli4ct}
\end{figure}

\begin{figure*}[t]
 \centering
 \includegraphics[width=0.91\textwidth]{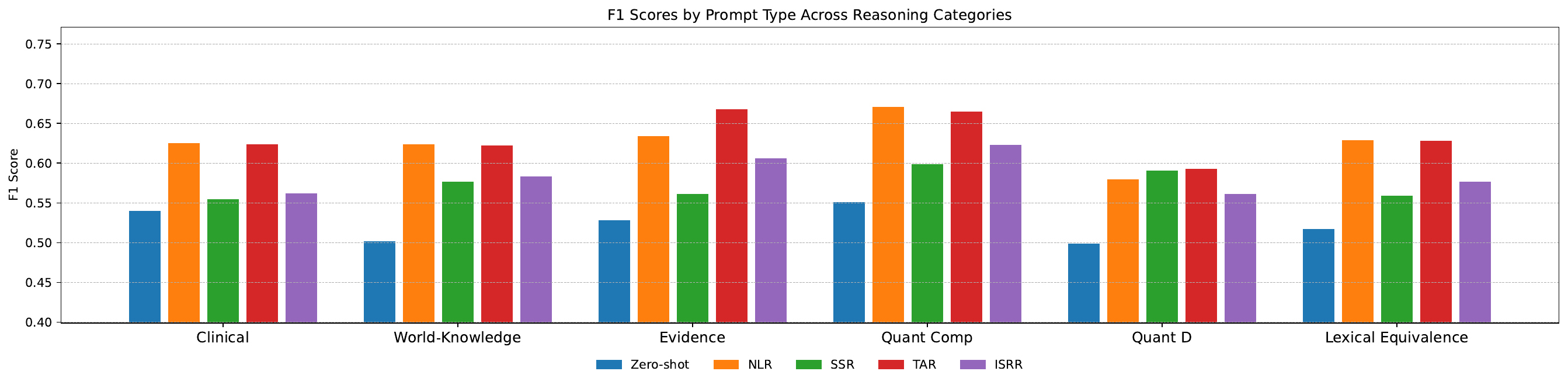}
 \caption{Performance by Reasoning Type on the NLI4CT Test Set (Macro F\textsubscript{1})}
\label{fig:reasoning}
\end{figure*}

\section{Empirical Evaluation}
\subsection{Representative Prompting Strategies}
\label{sec:prompting}

We instantiate one representative prompt per abstract category defined in Table~\ref{tab:prompt_structure}: Chain-of-Thought (NLR), Self-Critique (ISRR), ReACT (TAR), and QuaSAR (SSR), along with a zero-shot baseline. Each prompt reflects the structural pattern and reasoning mode characteristic of its respective category. Full prompt definitions, templates, and examples are provided in the Appendix, with representative reasoning trajectories illustrated in Figure~\ref{fig:main}.

\subsection{Dataset Selection}

\paragraph{NLI4CT}The NLI4CT dataset~\cite{jullien2023nli4ct}, defines an NLI task over structured sections of clinical trial reports (CTRs) from \texttt{ClinicalTrials.gov}. Given a natural language \textit{statement}, and a corresponding \textit{CTR premise}, a passage drawn from one of four sections: \textit{Eligibility}, \textit{Intervention}, \textit{Results}, or \textit{Adverse Events}, predict whether the premise entails or contradicts the statement by assigning one of two labels: \textit{Entailment} or \textit{Contradiction}. Each NLI4CT test instance was expert-annotated with one or more of six reasoning categories, indicating the types of reasoning needed to solve it. Full details are provided in the Appendix.

\paragraph{Generalisation Datasets}
To evaluate cross-domain generalisation, we test on two external clinical NLI datasets: MedNLI \cite{romanov2018lessons} and the TREC 2022 Clinical Trials Track \cite{roberts2022overview}. MedNLI provides sentence-level entailment labels derived from MIMIC-III patient notes, while TREC frames eligibility inference over synthetic patient-trial pairs. We use balanced 400-instance subsets from each dataset, reformulated as binary NLI tasks. Full dataset details are provided in the Appendix.

\subsection{Demonstration Generation}
Demonstrations are generated using \textsc{GPT-4o-Mini} \cite{openai_gpt4o_mini_2024} on 1,900 instances from the combined NLI4CT training and development sets for each prompt type. Only correctly classified instances are retained to construct high-precision demonstration sets (Table~\ref{tab:democount}).

\subsection{Model Selection}
\label{sec:model-selection}

We evaluate four compact, instruction-tuned language models: \texttt{LLaMA-3.2-3B} \cite{grattafiori2024llama},  \texttt{Qwen-2.5-3B-Instruct} \cite{qwen2},  \texttt{DeepSeek-R1-Distill-Qwen-1.5B} \cite{deepseekai2025deepseekr1incentivizingreasoningcapability}, and  \texttt{Phi-4–Mini-Reasoning-3.8B} \cite{xu2025phi}. This selection of smaller models reflects the practical constraints of real-world clinical settings. Regulatory and infrastructure limitations often necessitate on-premise deployment, where sub-4B models offer a tractable trade-off between computational cost, memory footprint, and task performance. Full details can be found in the Appendix.

\subsection{Fine-Tuning Details}
Each model is fine-tuned with LoRA using 500 randomly selected demonstrations corresponding to each prompt type. A uniform hyperparameter configuration is applied across all models to ensure consistency, with the exception of Phi-4, which requires a distinct target module configuration due to architectural constraints. Full adapter specifications and training details are provided in the Appendix.

\section{ Results on NLI4CT}

We present the results of the main experiments on the NLI4CT dataset in Figure 2, 3, and 4 for different prompting strategies, fine-tuning, and model configuration. 

\paragraph{Results Overview}
Overall, the results on NLI4CT reveal three central factors shaping performance on clinical NLI: prompt strategy, reasoning type sensitivity, and LoRA-based fine-tuning.
Controlling for model identity and LoRA status, prompt strategy explains \textbf{44\,\%} of the variance in macro–$F_{1}$ on NLI4CT and \textbf{30--44\,\%} within each individual reasoning class. Regarding the prompting strategy, we found that NLR and TAR yield the highest average $F_{1}$ scores. However, different prompts induce distinctive precision–recall profiles: ISRR maximises recall, whereas SSR maximises precision. Additionally, the results indicate that prompt strategies redistribute model strengths and weaknesses across reasoning types, rather than yielding uniform performance gains.
Regarding the inference types, we found that textit{Evidence} and \textit{Quantitative Comparison} are the least challenging (mean $F_{1}\,\geq\,0.653$), whereas \textit{Quantitative Derivation} remains the most difficult, exhibiting strong dependence on model size and limited benefit from LoRA. 
Finally, we observe that LoRA fine-tuning consistently improves macro–$F_{1}$ by \textbf{+8–12} \% for every prompt and model type except zero-shot, increasing also answer validity and format alignment above \textbf{97\,\%}.

\subsection{Prompting Strategy and Reasoning Types}

In this section, we analyse in more detail the performance and impact of the prompting strategies across different reasoning types on NLI4CT. 
The results are shown in Figure~ \ref{fig:eta}, \ref{fig:nli4ct} and ~\ref{fig:reasoning}, Tables~\ref{tab:anova-global-effects}, \ref{tab:per-reasoning-anova}, \ref{tab:lift-analysis}, \ref{tab:metrics} and~\ref{tab:normal-vs-base-summary}. 

\paragraph{Prompt strategy explains 44\% of the variance in F$_1$ performance, after controlling for the effects of model identity and LoRA.} Using a fixed-effects Type II ANOVA controlled for model architecture and LoRA adaptation (shown in Table~\ref{tab:anova-global-effects}, Figure~\ref{fig:eta}). Prompt strategy captures 44 \% of the variance left unexplained by the other two factors ($\eta^2_\text{partial}=0.440$, $p<0.001$). This effect is statistically significant and comparable to model architecture (51\%) and greater than LoRA (36\%). This confirms that prompt structure is a primary lever on performance.

\paragraph{NLR and TAR achieve the highest average F\textsubscript{1} across all models, regardless of size or fine-tuning} Across all model groups, and LoRA configurations, prompts rank consistently by average F\textsubscript{1} score. Specifically, NLR, TAR, ISRR, SSR, and Zero-shot, with the only deviation occurring in the base small model setting, where Zero-shot slightly outperforms SSR (Figure~\ref{fig:nli4ct}). This consistency suggests that the effectiveness of prompt methods is invariant to both model scale and LoRA. SSR and ISSR yield the lowest macro-$F_{1}$ scores in the base setting (Figure~\ref{fig:nli4ct}), though LoRA narrows the gap. This reflects the fact that SSR and ISSR impose greater planning overhead, longer-range dependency tracking, and tighter output constraints than NLR or TAR.

\paragraph{Prompt strategies enable precision–recall trade-offs}
As seen in Table~\ref{tab:metrics}, ISRR on GPT-4o-mini maximises recall (0.890) but sacrifices precision (0.648), while SSR inverses the trend (precision 0.832, recall 0.727). This suggests that prompt choice enables a tunable precision–recall trade-off.

\paragraph{Prompt strategy explains 30\% to 44\% of variance in F\textsubscript{1} across reasoning types, when controlling for model architecture and LoRA}
Fixed-effects Type II ANOVA per reasoning type shows prompt strategy explains 30–44\% of F\textsubscript{1} variance ($\eta^2_\text{partial}=0.3$–0.435, all $p<0.022$), independent of model and LoRA (Table~\ref{tab:per-reasoning-anova}, Figure~\ref{fig:eta}). For World Knowledge and Quantitative Comparison, prompt design explained the largest share of variance. The average variance explained by prompt strategy within individual reasoning types is 35\% compared to 44\% overall, indicating that prompts reshape the distribution of strengths and weaknesses across reasoning types, rather than providing consistent gains overall.

\paragraph{Zero-shot performance confirms task difficulty and impact of adaptation} Zero-shot F\textsubscript{1} scores for small models peak at 0.589 and average just 0.535, while GPT-4o-mini achieves only 0.643 (Figure~\ref{fig:nli4ct}). These results confirm that NLI4CT is a non-trivial task. Prompts that incorporate reasoning scaffolds, such as NLR and TAR consistently improve performance, with gains of up to +14 F\textsubscript{1} over the strongest zero-shot baselines.

\paragraph{NLR has the highest overall F\textsubscript{1}, but TAR Leads on Evidence and Quantitative Derivation} NLR achieves the highest mean F\textsubscript{1} across all reasoning classes (\textbf{0.660}, Table~\ref{tab:lift-analysis}, Figure~\ref{fig:reasoning}), only outperformed by TAR in the \textit{Evidence} and \textit{Quantitative Derivation} classes. While NLR outperforms most alternatives, no single strategy is currently optimal across all reasoning types.

\paragraph{Evidence and quantitative comparison represent the least challenging reasoning types (F\textsubscript{1} $\geq$ 0.653)} Across all models and prompting methods, \textit{Quantitative Comparison} (\textbf{F\textsubscript{1} = 0.663}) and \textit{Evidence} (\textbf{F\textsubscript{1} = 0.653}) emerge as the most tractable categories (Table~\ref{tab:normal-vs-base-summary}, (Figure~\ref{fig:reasoning})). These tasks likely benefit from simpler entailment structures or more direct linguistic cues. In contrast, \textit{Quantitative Derivation} remains the hardest, with a lower average F\textsubscript{1} and smaller variance in response to prompt or model changes.

\paragraph{SSR preferentially enhances quantitative reasoning}
SSR delivers the largest relative improvements on the two quantitative classes (Table~\ref{tab:lift-analysis}). 
With LoRA adaptation, the mean macro-$F_{1}$ for \textit{Quantitative Comparison} rises from $0.530$ to $0.668$ (\textbf{+$0.138$}, +26 \%), and for \textit{Quantitative Derivation} from $0.553$ to $0.628$ (\textbf{+$0.075$}, +14.7 \%). No other prompt attains comparable fractional lift on these categories. While designed for formal semantic clarity, SSR's predicate logic structure appears well aligned with quantitative reasoning, likely due to its explicit decomposition of claims into tractable, logic-based subcomponents.

\paragraph{Practical Implications}
These findings establish prompt design as a key determinant of performance in clinical natural language inference, with effects comparable to or exceeding those of model architecture and parameter-efficient adaptation. While NLR and TAR consistently yield the highest overall performance, no single prompt strategy optimally addresses all reasoning types. Prompt choice enables systematic trade-offs between precision and recall and reshapes the distribution of reasoning capabilities, with certain strategies conferring selective advantages on specific reasoning categories. Moreover, substantial variation in task difficulty across reasoning types suggests that prompt design should be aligned with the inferential demands of the target application. Future work in clinical NLI should prioritise the identification of dominant reasoning types and the development of prompt strategies tailored to their specific requirements.

\begin{figure}[t]
 \centering
 \includegraphics[width=\columnwidth]{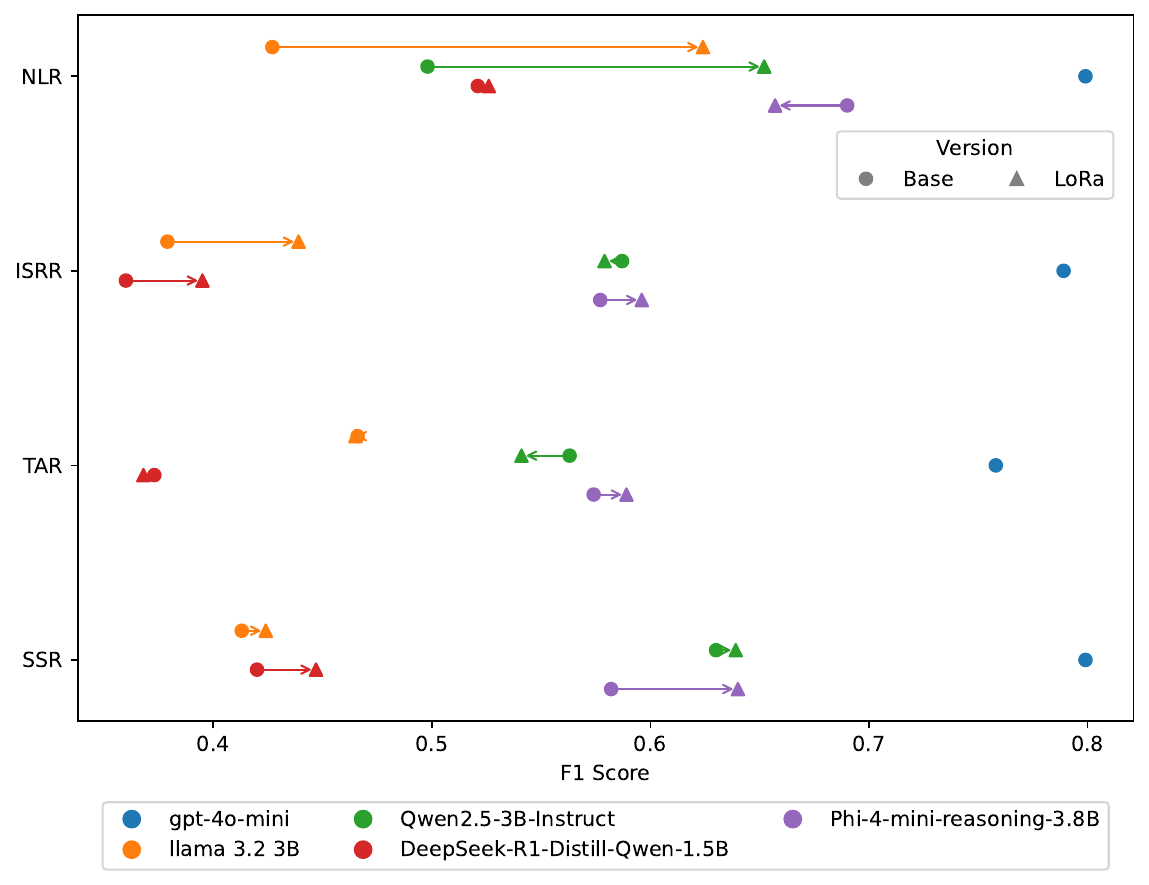}
 \caption{Macro F\textsubscript{1} Scores on the MedNLI Test Set}
\label{fig:mednli}
\end{figure}

\begin{figure}[t]
 \centering
 \includegraphics[width=\columnwidth]{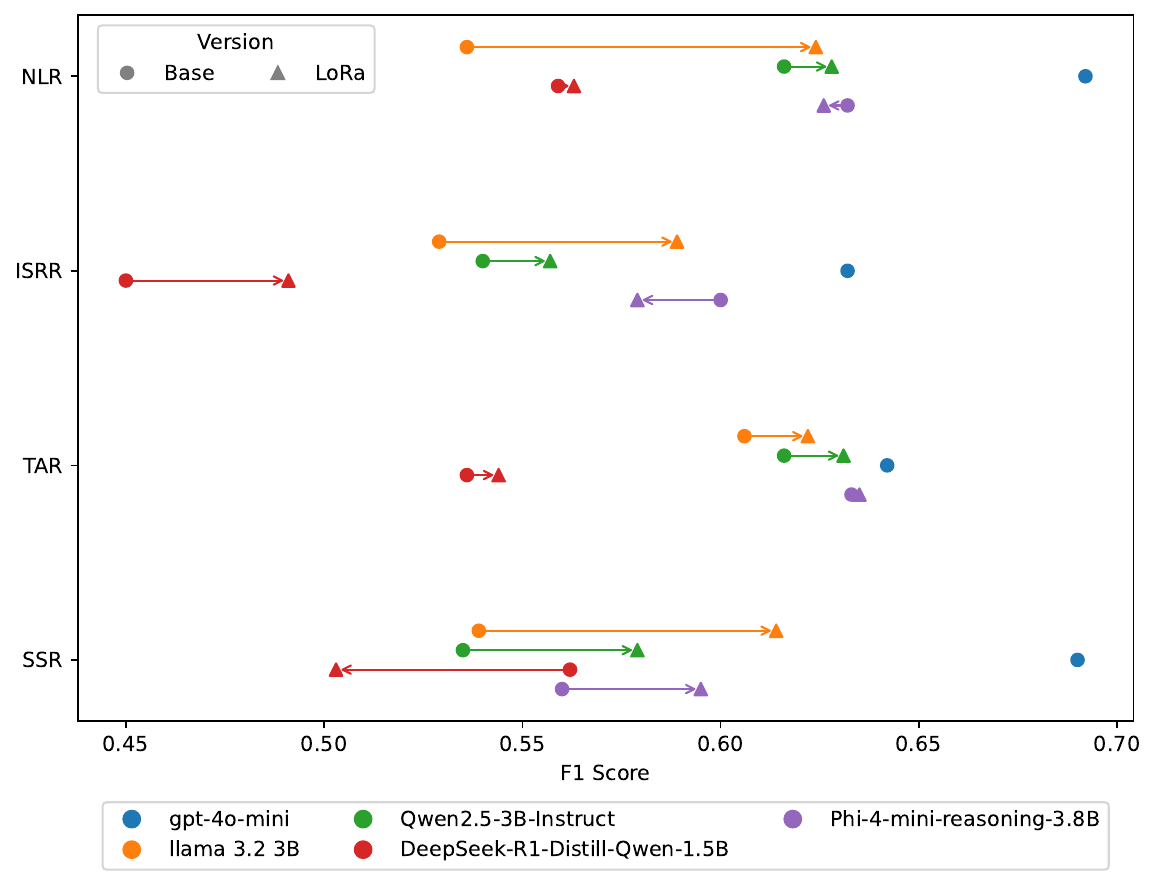}
 \caption{Macro F\textsubscript{1} Scores on the TREC Test Set}
\label{fig:trec}
\end{figure}

\subsection{The Impact of LoRA Fine-Tuning}

In this section, we analyse the impact of LoRA fine-tuning. Results are presented in Figures~\ref{fig:nli4ct} and~\ref{fig:lift}, Tables~\ref{tab:lift-analysis} and~\ref{tab:metrics}.

\paragraph{LoRA fine-tuning improves F\textsubscript{1} by +8–12\%for All Prompts Except Zero-Shot}
On average LoRA yields substantial gains in F\textsubscript{1} performance of +8-12\% for all prompts, except Zero-Shot that loses 2 \% (Figure~\ref{fig:nli4ct}) due to the absence of intermediate supervision signals. The highest relative F\textsubscript{1} gain from LoRA observed with SSR on LLaMA 3.2 3B with a +44.7\% relative improvement (Figure~\ref{fig:nli4ct}). This disparity between Zero-Shot and SSR highlights a synergy between fine-tuning and prompt structure: parameter adaptation is most beneficial when paired with prompts that elicit interpretable reasoning trajectories.

\paragraph{LoRA fine-Tuning raises answer validity to over 97\% across all models}
Validity is defined as a single, correctly formatted label from the task-defined set (i.e., \textit{entailment} or \textit{contradiction}). LoRA-tuned models generate well-formed outputs, exceeding 97\% validity across all configurations (Table~\ref{tab:metrics}). In contrast, base models are prone to formatting errors or incomplete responses, for instance, DeepSeek-base with ISRR produces valid outputs only 70.6\% of the time.


\paragraph{LoRA most benefits structured semantic reasoning, but struggles with layered arithmetic reasoning} LoRA improves F\textsubscript{1} on all reasoning types, with the largest gains in structured semantic reasoning (Figure~\ref{fig:lift}). Averaged across all prompt methods, largest F\textsubscript{1} gains are \textit{Evidence} (+0.117), \textit{Quantitative Comparison} (+0.096), and \textit{Lexical Equivalence} (+0.092) (Table~\ref{tab:lift-analysis}). By contrast, \textit{Quantitative Derivation}, which saw the lowest lift (+0.047) requires additional layers of symbolic reasoning not easily optimized through text-based supervision \citep{nye2021show, wei2022chain}. These results imply that LoRA is more effective at reinforcing shallow symbolic patterns than deeper computational abstractions \citep{hu2022lora, workrethinking}.

\paragraph{Model size still affects F\textsubscript{1}, but best small model trails GPT-4o-mini by only 7.1\%} GPT-4o-mini maintains a consistent performance lead across all prompts, achieving a peak F\textsubscript{1} score of 0.8 with NLR (Figure~\ref{fig:nli4ct}). However, Phi-4-LoRA with NLR, reaches an F\textsubscript{1} of 0.729, within 7.1\% of GPT-4o-mini, despite a significantly lower parameter count and computational requirements.


\paragraph{Practical Implications}
LoRA offers an efficient and scalable approach to enhancing clinical NLI performance. Consistent F\textsubscript{1} improvements across reasoning types and validity rates exceeding 97\% indicate that LoRA improves both predictive accuracy and output reliability. Moreover, prompt-based LoRA tuning enables compact models such as Phi-4 to achieve performance approaching that of frontier systems like GPT-4o-mini, offering a cost-effective alternative. Nonetheless, the comparatively modest gains on tasks requiring layered quantitative reasoning highlight the limitations of text-only supervision and the potential need for symbolic or structured augmentation.

\section{Generalisation to Different NLI Tasks}

In this section, we analyse the results of the generalisation study performed on MedNLI and TREC (training only on NLI4CT). The results are presented in Figures ~\ref{fig:mednli} and \ref{fig:trec} and  Tables~\ref{tab:med_metrics}, and \ref{tab:trec_metrics}.

\paragraph{LoRA improves F\textsubscript{1} in 75\% of cases (24/32), with gains up to +20\%}
Across the MedNLI and TREC datasets, LoRA-tuning on NLI4CT improves F\textsubscript{1} in 24/32 model–prompt combinations (Figures~\ref{fig:mednli}, \ref{fig:trec}, Table~\ref{tab:med_metrics}, \ref{tab:trec_metrics}). Gains range from increases of +0.01 to +0.20, with no degradation exceeding –0.03, with no drop exceeding –0.03, indicating that LoRA tuning is generally safe and beneficial for improving generalisation across model–prompt pairs.

\paragraph{LoRA preserves structural validity on different tasks}
LoRA is capable of enforcing output structure and reliability on tasks where the models are not explicitly trained on. All LoRA-tuned models produce 100\% valid outputs on MedNLI and $\geq$94.5\% on TREC. While F\textsubscript{1} increases are mostly recall-driven (e.g., Qwen2.5-NLR recall: 0.667 to 0.725), precision generally stays flat or declines slightly. This suggests that LoRA contributes to recovering more true positives by regularizing output format (Figures ~\ref{fig:mednli} and~\ref{fig:trec}).

\paragraph{LoRA reduces the gap between models on complex tasks.} Interestingly, we found that LoRA contribute to substantially reducing the gap between the fine-tuned model and frontier models on TREC, the task that is generally the most challenging for all the tested models. As shown in Figure 6, even if not explicitly trained on TREC, smaller models equipped with LoRA fine-tuning on NLI4CT can achieve performances that are comparable with GPT-4o-mini (especially using the TAR prompting strategy).

\paragraph{Practical Implications}
LoRA tuning on NLI4CT produces models that generalise reliably across prompt and model variants, and clinical NLI tasks with consistent gains in recall and output validity. This indicates the potential of parameter-efficient adaptation strategies to effectively deploy open-source and smaller language models on highly specialised domains for practical applications.


\section{Related Work}

\paragraph{Prompting‐Driven Reasoning in Clinical settings} Prompt engineering has substantially improved LLM reasoning, through techniques such as Chain‐of‐Thought (CoT) ~\cite{wei2022chain}, self‐consistency decoding~\cite{wang2022self}, TAR’s iterative verification loops~\cite{yao2023react}, and self‐critique refinement~\cite{madaan2023self} have each improved performance on general reasoning benchmarks. 
In clinical NLP, Sci‐CoT distils GPT–4 reasoning traces into compact models for scientific question answering~\cite{ma2023sci}, while MedPrompt steers GPT–4 with few‐shot CoT and self‐generated explanations, surpassing domain‐specific systems on nine medical challenge sets~\cite{nori2023can}. 
Despite these advances, systematic head-to-head comparisons of multiple prompting paradigms remain rare—especially for tasks such as clinical-trial NLI that require a blend of clinical knowledge, commonsense inference, and quantitative reasoning.

\paragraph{Prompting Strategies for NLI4CT} Across two SemEval shared tasks~\cite{jullien2023semeval,jullien2024semeval}, participants explored various prompting techniques for NLI4CT: zero-shot templates, few-shot in-context learning, CoT, contrastive CoT, and retrieval-augmented prompting. However, these strategies were trialled on heterogeneous base models, obscuring the extent to which observed gains originate from prompt design. Additionally, meta-analysis indicates that CoT offers the greatest benefit for mathematical and symbolic tasks~\cite{sprague2024cot}, but contributes less to the commonsense, knowledge-based, and pragmatic reasoning demanded by clinical NLI.

\paragraph{Cross-Domain Insights.}
Recent cross-domain studies have begun to isolate the effect of prompting on distinct reasoning categories (Table~\ref{tab:cross_reasoning_studies}). 
ThinkPatterns-21k~\cite{wen2025thinkpatterns}, Auto-CoT~\cite{zhang2022automatic}, and Self-Ask~\cite{press2022measuring}, for example, contrast several prompting patterns on mathematics, commonsense, and symbolic problems, while Beyond Accuracy~\cite{mondorf2024beyond} and Systematic Relational Reasoning (SRR) ~\cite{khalid2025benchmarking} probe logical, causal, and spatial reasoning. 
Yet none of these efforts considers clinical-trial NLI.

\paragraph{Limitations of Prior Work}
Prior work has shown the potential of prompt-based reasoning in clinical NLP and surveyed prompt strategies across domains. What remains lacking is a controlled evaluation of how these strategies transfer specifically to clinical-trial NLI.

\section{Conclusion}\label{sec:conclusion}

We present the first controlled study of prompting and parameter-efficient adaptation for clinical-trial NLI.
\begin{itemize}\setlength\itemsep{2pt}
\item \textbf{Prompting paradigms \& reasoning categories.} We formalise four prompting paradigms and six clinically grounded reasoning categories, enabling systematic prompt–reasoning analysis, and dataset annotation.
\item \textbf{Prompt structure is a primary driver performance.}  When controlling for model architecture and adaptation, prompt structure explains up to 44 \% of the macro-F\textsubscript{1} variance on NLI4CT.
\item \textbf{LoRA lifts compact models.} On 3–4 B models, LoRA adds 8–12\% macro-F\textsubscript{1}, raises answer validity above 97 \%, and narrows the gap to GPT-4o-mini to 7.1 \%.
\item \textbf{Reasoning-aware evaluation exposes trade-offs.} Prompt choice accounts for 30–44 \% of the variation across reasoning categories.
\item \textbf{Generalisation across clinical NLI benchmarks.} LoRA improves F\textsubscript{1} in 75 \% of model–prompt pairs on MedNLI and TREC-CT, showing robust generalisation.
\end{itemize}

Code, annotations, prompts, demonstrations, and checkpoints are released to foster reproducible research on principled prompting and efficient adaptation for trustworthy clinical reasoning.

\bibliography{custom}

\appendix

\section{Appendix}
\label{sec:appendix}

\subsection{Limitations}
\label{sec:limitations}

Despite the encouraging gains reported, several factors constrain the scope and
generalizability of our findings.

\paragraph{Synthetic Demonstrations.}
All demonstrations were generated with
\textsc{GPT-4o-Mini}. This risks
importing factual or stylistic biases present in the teacher model into the
fine-tuned students. Future work should explore human-curated or adversarially
filtered demonstration pools to mitigate inherited bias.

\paragraph{Clinical Readiness.}
The models are trained and evaluated solely on
benchmark datasets and have not undergone validation on real patient data,
error-sensitivity analyses, or regulatory review. Consequently, these systems
are \emph{not} suitable for clinical use. Additionally, deploying NLI models in healthcare raises additional ethical duties, including safeguarding patient privacy, preventing harmful
automation bias, and ensuring transparency of decision support. These issues
are beyond the scope of this study.

\paragraph{Model Scale.}
We restrict our experiments to checkpoints below
4 B parameters. Larger models may interact differently with the same prompt
strategies, potentially altering the utility of LoRA, or certain prompt strategies.

\paragraph{Annotation Reliability.}
Reasoning-type labels for NLI4CT were
provided by a single domain expert. The absence of inter-annotator agreement
measures limits our ability to quantify subjectivity in the error analysis.

\paragraph{Uniform Training Protocol.}
A fixed set of hyper-parameters is applied
to all model–prompt combinations for consistency. While this controls for
tuning variance, it may inadvertently favour architectures/prompt strategies whose inductive
biases align better with the chosen configuration.

\subsection{Model Links}
\label{sec:model-links}

The following publicly available models were used in this study:

\begin{itemize}
 \item \textbf{LLaMA-3.2–3B}: \url{https://huggingface.co/meta-llama/Llama-3.2-3B-Instruct}
 \item \textbf{Qwen-2.5–3B-Instruct}: \url{https://huggingface.co/Qwen/Qwen2.5-3B-Instruct}
 \item \textbf{DeepSeek–R1–Distill–Qwen–1.5B}: \url{https://huggingface.co/deepseek-ai/DeepSeek-R1-Distill-Qwen-1.5B}
 \item \textbf{Phi-4–Mini–Reasoning–3.8B}: \url{https://huggingface.co/microsoft/Phi-4-mini-reasoning}
\end{itemize}

\subsection{Training Configuration}
\label{sec:training}

We fine-tuned our model using the Hugging Face \texttt{Trainer} API in conjunction with parameter-efficient fine-tuning (PEFT) via LoRA~\cite{hu2022lora}. Below, we detail the hyperparameters and training configuration used in our experiments.

\subsubsection{LoRA Configuration}
We applied LoRA to the attention and feed-forward layers of the model with the following settings:
\begin{itemize}
 \item \textbf{Task type:} Causal Language Modeling (CAUSAL\ LM)
 \item \textbf{LoRA rank ($r$):} 8
 \item \textbf{LoRA scaling factor ($\alpha$):} 16
 \item \textbf{LoRA dropout:} 0.1
 \item \textbf{Target modules:} \texttt{[q\_proj, k\_proj, v\_proj, o\_proj, gate\_proj, up\_proj, down\_proj]}
 \item \textbf{Target modules for Phi:} \texttt{['down\_proj', 'gate\_up\_proj', 'qkv\_proj', 'o\_proj']}
 \item \textbf{Bias:} None
\end{itemize}

\subsubsection{Training Hyperparameters}
The training was performed using the following hyperparameter settings:
\begin{itemize}
 \item \textbf{Maximum sequence length:} 4096
 \item \textbf{Train batch size (per device):} 1
 \item \textbf{Eval batch size (per device):} 1
 \item \textbf{Gradient accumulation steps:} 16
 \item \textbf{Learning rate:} $2 \times 10^{-5}$
 \item \textbf{Weight decay:} 0.01
 \item \textbf{Warm-up steps:} 10
 \item \textbf{Maximum steps:} 500
 \item \textbf{Early stopping patience:} 2 evaluations
\end{itemize}

\subsubsection{Optimization}
The model selection criterion was the lowest evaluation loss on the validation set.

\FloatBarrier
\subsection{Prompt Descriptions \& Examples}
\label{sec:prompt-details}
\paragraph{NLR: Chain–of–Thought (CoT).}
CoT instructs the model to verbalise step-by-step reasoning, breaking down complex problems into more manageable parts, and has been shown to improve performance over standard prompting methods, particularly in domains that require multi-step reasoning, such as mathematical problem-solving, common-sense inference, and logical deduction \cite{wei2022chain}. See Table~\ref{tab:cot_ex} and Figure~\ref{fig:main} for an example of CoT reasoning on NLI4CT.

\paragraph{ISRR: Self‑Critic.}
The Self‑Critic prompt \cite{madaan2023self} embeds an explicit reflection phase: the model drafts an answer, critiques its own reasoning, and then revises accordingly. This three‑step workflow, exemplified in Table~\ref{tab:scrit_ex} and Figure~\ref{fig:main}, fosters error detection, reduces reasoning shortcuts, and has been linked to improved factual accuracy and safety in sensitive domains.

\paragraph{TAR: REACT (Reason \& Act).}
REACT \cite{yao2023react} alternates internal reasoning tokens (\textsc{Thought}) with explicit operations on the environment (\textsc{Action}) followed by feedback (\textsc{Observation}), as shown in Table~\ref{tab:TAR_ex} and Figure~\ref{fig:main}. In this implementation, \textsc{Action} acts as a self‑check that compares provisional inferences with the premise text. This iterative verify‑before‑commit loop has been shown to curb hallucinations in knowledge‑intensive tasks and to encourage the retrieval of supporting evidence.

\paragraph{SSR: Quasi–Symbolic Abstract Reasoning (QuaSAR).}
QuaSAR \cite{ranaldi2025improving} decomposes a problem into four quasi‑symbolic stages designed to tighten the model’s reasoning trajectory: (i) Abstraction: identify predicates, variables, and constants; (ii) Formalisation: Using the abstracted variables, formalise the logical structure and constraints of the problem(s); (iii) Explanation: derive the solution over the formal representation; and (iv) Answering: translate the result back into natural language. These steps are demonstrated in Table~\ref{tab:qs_ex} and Figure~\ref{fig:main}. This prompt enables explanatory arguments to be abstracted from specific world knowledge—for example, rather than learning 'The patient takes metformin, so they have diabetes,' the model learns the general pattern \textit{"Taking [Drug X] implies [Condition Y]"} supporting generalisation across cases \cite{ranaldi2025improving}.

\paragraph{Zero-Shot}
In the zero-shot setting, the model is prompted to produce a label directly, without any intermediate reasoning. This serves as a baseline or sanity check against which more structured strategies can be compared.

\begin{table}[H]
\centering
\scriptsize
\begin{tabular}{>{\ttfamily\raggedright\arraybackslash}p{0.9\columnwidth}}
\toprule
\textbf{TAR Instructions}
\\\midrule
\textbf{Statement:} Patients with a Vitamin D deficiency are excluded\\
\midrule
\textbf{Premise:} Inclusion Criteria..[46 words].. Serum 25OHD levels <40 ng/ml\\
\midrule
\textbf{Though:} I need to examine whether the inclusion criteria mentions Vitamin D\\
\midrule
\textbf{Action:} Check for any mention of vitamin D deficiency\\
\midrule
\textbf{Observation:} The criteria do not mention vitamin d deficiency directly. however, ..[13 words]..\\
\midrule
\textbf{Though:} The claim in the statement is supported by the premise. \textbf{Output: Contradiction}
\\
\bottomrule
\end{tabular}
\caption{Examples of TAR reasoning}
\label{tab:TAR_ex}
\end{table}

\begin{table}[H]
\centering
\scriptsize
\begin{tabular}{>{\ttfamily\raggedright\arraybackslash}p{0.9\columnwidth}}
\toprule
\textbf{SSR Instructions} \\
\midrule
\textbf{Statement:} Patients with a Vitamin D deficiency are excluded \\
\midrule
\textbf{Premise:} Inclusion Criteria..[46 words].. Serum 25OHD levels <40 ng/ml \\
\midrule
\textbf{Abstraction (s1):} Predicates: excluded(trial, condition) \par Variables: vitamin D deficiency ..[23 words] \\
\midrule
\textbf{Formalisation (s2):} Let d be the predicate indicating vitamin D deficiency. Let t$_1$ be the primary trial. We assert e$_1$: excluded\_from\_primary(t$_1$, d)..[48 words] \\
\midrule
\textbf{Explanation (s3):} Step 1: Assess the criteria given the condition of Vitamin D deficiency \par Step 2: Determine if Vitamin D..[28 words].. \\
\midrule
\textbf{Answering (s4):} \textbf{Output: Contradiction} \\
\bottomrule
\end{tabular}
\caption{Example of SSR reasoning}
\label{tab:qs_ex}
\end{table}

\begin{table}[H]
\centering
\scriptsize
\begin{tabular}{>{\ttfamily\raggedright\arraybackslash}p{0.9\columnwidth}} 
 \toprule
 \textbf{Chain of Though Instructions}
 \\\midrule
 \textbf{Statement:} Patients with a Vitamin D deficiency are excluded\\\midrule
 \textbf{Premise:} Inclusion Criteria..[46 words].. Serum 25OHD levels <40 ng/ml\\\midrule
\textbf{Reasoning:} Let's examine the premises. The inclusion criteria state that participants must have serum 25OHD levels$ < $40 ng/ml. Since Vitamin D deficiency is typically defined as
serum 25OHD$ < $40 ng/ml, this condition qualifies patients for inclusion, not exclusion. Therefore, the statement contradicts the premise.
\textbf{Output: Contradiction}\\
 \bottomrule
\end{tabular}
\caption{Example of CoT reasoning}
\label{tab:cot_ex}
\end{table}

\begin{table}[H]
\centering
\scriptsize
\begin{tabular}{>{\ttfamily\raggedright\arraybackslash}p{0.9\columnwidth}}
\toprule
\textbf{ISRR Instructions}\\
\midrule
\textbf{Statement:} Patients with a Vitamin D deficiency are excluded\\
\midrule
\textbf{Premise:} Inclusion Criteria..[46 words].. Serum 25OHD levels <40 ng/ml\\
\midrule
\textbf{Draft:} The statement claims that patients with a vitamin d deficiency are excluded from the primary trial ..[41 words].. 
\textbf{Output: Entailment}\\
\midrule
\textbf{Critical Comments:} The draft fails to identify the link between Serum 250HD and Vitamin D ..[52 words].. Additionally, the terminology for excluded should be consistent .\\
\midrule
\textbf{Final Response:} The inclusion criteria state that participants must have serum 25OHD levels$ < $40 ng/ml ..[37 words]..
\textbf{Output: Contradiction} \\
\bottomrule
\end{tabular}
\caption{Examples of ISRR reasoning}
\label{tab:scrit_ex}
\end{table}

\begin{table}[H]
\centering
\scriptsize
\begin{tabular}{>{\ttfamily\raggedright\arraybackslash}p{0.9\columnwidth}} 
 \toprule
 \textbf{Baseline Prompt}
 \\\midrule
 Determine whether the statement is an entailment or a contradiction of the premise. Answer only with either "entailment" or "contradiction".\\
 \textbf{Statement:} \{statement\}\\
 \textbf{Premise:} \{premise\}\\
 \textbf{Output:}
 \\\bottomrule
\end{tabular}
\caption{Baseline prompt for NLI4CT}
\label{tab:baseline_prompt}
\end{table}

\begin{table}[H]
\centering
\scriptsize
\begin{tabular}{>{\ttfamily\raggedright\arraybackslash}p{0.9\columnwidth}} 
 \toprule
 \textbf{Chain of Thought Prompt}
 \\\midrule
 You are given a premise and a statement. Your task is to determine the relationship between the statement and the premise by analyzing them carefully.

Instructions:

• Carefully read the statement and the premise. 
• Think through your reasoning step by step, considering whether the statement logically follows from the premise or contradicts it. 
• Use chain-of-thought reasoning to reach your conclusion. 
• Based on your analysis, determine whether the statement is an 'entailment' or a 'contradiction' of the premise. 
• In your response, first provide your chain-of-thought reasoning, and then output 'entailment' or 'contradiction' as your final answer. 
• You must end your response with "output: entailment" or "output: contradiction".\\
\textbf{Statement:} \{statement\}\\
\textbf{Premise:} \{premise\}\\
\textbf{Reasoning:}
\\\bottomrule
\end{tabular}
\caption{Chain-of-thought prompt for NLI4CT}
\label{tab:cot_prompt}
\end{table}

\begin{table}[H]
\centering
\scriptsize
\begin{tabular}{>{\ttfamily\raggedright\arraybackslash}p{0.9\columnwidth}} 
 \toprule
 \textbf{QuaSAR Prompt}
 \\\midrule
\#Role 
You are an experienced expert skilled in answering complex problems through logical reasoning and structured analysis.

\#Task 
You are presented with an entailment problem that requires logical reasoning and systematic problem-solving. Given a statement and a premise, you are required to determine whether the statement follows from the premise. If the statement follows from the premise, end your response with "output: entailment". If the statement contradicts the premise, end your response with "output: contradiction". Please determine the entailment following these steps rigorously.

\#Steps 
1) Please consider the following statement and premise and exemplify the relevant predicates, variables, and constants. Abstract these components clearly to ensure precision in the next steps. Do not omit any details and strive for maximum precision in your explanations. Refer to this step as Abstraction (s1)

2) For each predicate, variable and constant defined in s1, translate the statement and premise in formal symbolic representation. Please ensure that the formalisation captures the logical structure and constraints of the statement and premise. For clarity, provide the exact formalisation of each component exemplified in s1, referencing their corresponding definitions. Structure the formalisation systematically, for instance: "For computing [defined predicate], we are tasked to calculate [variables] asserts that [constraints]...". Refer to this step as Formalisation (s2)

3) Please consider the formalisation in s2 in detail, ensure this is correct and determine the entailment by breaking down the steps operating a symbolic representation. Combine variables, constants, and logical rules systematically at each step to find the solution. For clarity, provide clear reasoning for each step. Structure the explanation systematically, for instance: "Step 1: Calculate... Step 2:....". Refer to this step as Explanation (s3)

4) In conclusion, behind explaining the steps supporting the final answer to facilitate the final evaluation, extract the answer in a short and concise format by marking it as "output: entailment/contradiction" At this stage be strict and concise and refer to this step as Answering (s4).\\\midrule

\textbf{Statement:} \{statement\}\\
\textbf{Premise:} \{premise\}\\
\textbf{Abstraction (s1):}
\\\bottomrule
\end{tabular}
\caption{SSR prompt for NLI4CT}
\label{tab:formal_prompt}
\end{table}

\begin{table}[H]
\centering
\scriptsize
\begin{tabular}{>{\ttfamily\raggedright\arraybackslash}p{0.9\columnwidth}} 
 \toprule
 \textbf{REACT Prompt}
 \\\midrule
You are given a premise and a statement. Your task is to determine the relationship between the statement and the premise. 
Using thought, observation and action steps.

Instructions: 
- Begin with a \textbf{Thought}, where you explain your reasoning step by step. 
- Use \textbf{Action} to simulate operations like “Check if X is mentioned in the premise” or “Compare X with Y”. 
- Follow with \textbf{Observation} to simulate what you learned from that action. 
- Repeat this process until you reach a conclusion. 
- You must end your response with "output: entailment" or "output: contradiction".\\
\textbf{Statement:} \{statement\}\\
\textbf{Premise:} \{premise\}\\
\textbf{Reasoning:}
\\\bottomrule
\end{tabular}
\caption{REACT Prompt for NLI4CT}
\label{tab:tao_prompt}
\end{table}

\begin{table}[H]
\centering
\scriptsize
\begin{tabular}{>{\ttfamily\raggedright\arraybackslash}p{0.9\columnwidth}} 
 \toprule
 \textbf{Self-Critique Prompt}
 \\\midrule
You are given a premise and a statement. Your task is to determine the relationship between the statement and the premise. 
First generate a \textbf{Draft Response}, then generate \textbf{Critical Comments}, then generate a \textbf{Final Response}.

\textbf{Requirements:} 
1. \textbf{Draft Response:} Generate an initial response 
2. \textbf{Critical Comments:} 
Analyze your draft response by considering: 
- Potential weaknesses or gaps 
- Logical flaws or inconsistencies 
- Missing perspectives or alternatives 
- Areas for improvement 
- Suggestions for a better version 
- Steering toward the given answer 

The critical comments should: 
- Be specific and actionable 
- Reference particular parts of the draft 
- Suggest concrete improvements 
- Consider different angles or approaches 
- Guide towards a more comprehensive solution 

3. \textbf{Final Response:} Generate a final response that incorporates the critical comments. 

You must end your response with "output: entailment" or "output: contradiction".\\
\textbf{Statement:} \{statement\}\\
\textbf{Premise:} \{premise\}\\
\textbf{Draft Response:}
\\\bottomrule
\end{tabular}
\caption{ISRR prompt for NLI4CT}
\label{tab:self_critique_prompt}
\end{table}

\subsection{NLI4CT Description}
NLI4CT is designed to support real-world clinical tasks and enable comprehensive evaluation of clinical reasoning. It comprises 2,400 premise–statement pairs, split 70/20/10 across train, test, and development sets. Examples where manually curated and labeled for entailment, with CTR sections and labels evenly distributed across all splits. Test set instances are annotated with 1-4 reasoning types, with an average of 1.6 types per instance. The distribution of reasoning types is shown in Table~\ref{tab:reasoning_counts}.

\begin{table}[H]
\small
\centering
\begin{tabular}{p{4cm} p{3cm}}
\toprule
\textbf{Reasoning Type} & \textbf{Number of Instances} \\
\midrule
Lexical Equivalence & 223 \\
Clinical  & 131 \\
Quantitative Comparison & 148 \\
Quantitative Derivation & 95 \\
Evidence & 121 \\
World‐Knowledge Inference & 84 \\
\bottomrule
\end{tabular}
\caption{Number of instances requiring each type of reasoning in NLI4CT.}
\label{tab:reasoning_counts}
\end{table}

\FloatBarrier
\subsection{Generalisation Dataset Descriptions}
\label{sec:dataset-details}
\subsubsection{MedNLI}

MedNLI \cite{romanov2018lessons} is a clinical NLI dataset constructed from de-identified patient notes in the MIMIC-III database \cite{johnson2016mimic}. Each instance pairs a short premise—typically a single sentence from a clinical note—with a hypothesis written by a medical expert and labeled as \textit{Entailment}, \textit{Contradiction}, or \textit{Neutral}. For example, the premise \textit{“He denied headache or nausea or vomiting”} and hypothesis \textit{“He is afebrile”} is labeled \textit{Neutral}, since the temperature status is not entailed or contradicted by the premise. For evaluation, we sample a balanced subset of 400 examples from the 1,422-instance test set.

\subsubsection{TREC 2022 Clinical Trials Track}

The TREC 2022 Clinical Trials Track \cite{roberts2022overview} presents a patient-to-trial retrieval task. Each input is a synthetic patient case description (topic) typically 5--10 sentences in length—that simulates an admission statement in an electronic health record. The retrieval corpus consists of approximately 375k CTRs from \texttt{ClinicalTrials.gov}, of which a subset has been labeled as \textit{Eligible}, \textit{Excluded}, or \textit{Not Relevant}, relative to a given topic. 

For this study, the TREC task is reformulated as a binary NLI classification problem. A balanced dataset of 400 trial–topic pairs is constructed, evenly split between the \textit{Eligible} and \textit{Excluded} classes. The \textit{Not Relevant} category is omitted, as it primarily reflects retrieval failure rather than contradiction with eligibility criteria. An example instance is shown in Table~\ref{tab:trec_example}.

\begin{table}[H]
\centering
\small
\begin{tabular}{@{}p{1.5cm}p{5.5cm}@{}}
\toprule
\textbf{Field} & \textbf{Content} \\
\midrule
Topic & \textit{23-year-old man with exertional syncope, family history of sudden death, harsh systolic murmur, and septal hypertrophy.} \\\midrule
Trial Description & \textit{Diagnostic strategies for suspected pulmonary embolism in outpatients.} \\\midrule
Label &Excluded \\
\bottomrule
\end{tabular}
\caption{Example instance from the TREC dataset.}
\label{tab:trec_example}
\end{table}

\FloatBarrier
\subsection{Results Tables}

\begin{table*}[t]
\small
\centering
\begin{tabular}{lccccc}
\toprule
\textbf{Model} & \textbf{F\textsubscript{1}} & \textbf{Recall} & \textbf{Precision} & \textbf{Accuracy}& \textbf{\% Valid} \\
\midrule
\noalign{\vskip -3pt}
\multicolumn{5}{c}{\textbf{NLR}} \\
\noalign{\vskip -3pt}
\midrule
gpt-4o-mini & 0.692 & 0.700 & 0.722 & 0.699 & 1.000\\
llama 3.2 3B & 0.536 & 0.843 & 0.551 & 0.573 & 0.978 \\ 
DeepSeek-R1-Distill-Qwen-1.5B & 0.559 & 0.747 & 0.556 & 0.573 & 0.925 \\
Phi-4-mini-reasoning-3.8B & 0.632 & 0.588 & 0.640 & 0.633 & 0.980 \\
Qwen2.5-3B-Instruct & 0.616 & 0.667 & 0.610 & 0.618 & 0.968\\
Qwen2.5-3B-Instruct LoRa & 0.628 & 0.725 & 0.612 & 0.632 & 0.998 \\ 
llama 3.2 3B LoRa & 0.624 & 0.665 & 0.616 & 0.625 & 1.000\\
DeepSeek-R1-Distill-Qwen-1.5B LoRa & 0.563 & 0.775 & 0.560 & 0.580 & 0.995 \\
Phi-4-mini-reasoning-3.8B LoRa& 0.626 & 0.505 & 0.678 & 0.632 & 0.958 \\ 
\midrule
\noalign{\vskip -3pt}
\multicolumn{5}{c}{\textbf{SSR}} \\
\noalign{\vskip -3pt}
\midrule
gpt-4o-mini & 0.690 & 0.693 & 0.700 & 0.693 & 1.000\\
llama 3.2 3B & 0.539 & 0.650 & 0.537 & 0.544 & 0.882 \\ 
DeepSeek-R1-Distill-Qwen-1.5B &0.562 & 0.794 & 0.575 & 0.587 & 0.792 \\
Phi-4-mini-reasoning-3.8B & 0.560 & 0.497 & 0.573 & 0.561 & 0.980\\
Qwen2.5-3B-Instruct & 0.535 & 0.553 & 0.560 & 0.536 & 0.765\\
Qwen2.5-3B-Instruct LoRa & 0.579 & 0.599 & 0.576 & 0.579 & 0.985\\ 
llama 3.2 3B LoRa & 0.614 & 0.619 & 0.613 & 0.614 & 0.945\\
DeepSeek-R1-Distill-Qwen-1.5B LoRa & 0.503 & 0.697 & 0.516 & 0.519 & 0.920\\
Phi-4-mini-reasoning-3.8B LoRa& 0.595 & 0.464 & 0.634 & 0.603 & 0.907\\ 
\midrule
\noalign{\vskip -3pt}
\multicolumn{5}{c}{\textbf{TAR}} \\
\noalign{\vskip -3pt}
\midrule
gpt-4o-mini & 0.642 & 0.655 & 0.681 & 0.655 & 1.000\\
llama 3.2 3B & 0.606 & 0.775 & 0.585 & 0.616 & 0.963 \\ 
DeepSeek-R1-Distill-Qwen-1.5B &0.536 & 0.674 & 0.540 & 0.545 & 0.895 \\
Phi-4-mini-reasoning-3.8B & 0.633 & 0.574 & 0.659 & 0.634 & 0.963\\
Qwen2.5-3B-Instruct & 0.616 & 0.591 & 0.618 & 0.616 & 0.938 \\
Qwen2.5-3B-Instruct LoRa & 0.631 & 0.683 & 0.621 & 0.632 & 0.993\\ 
llama 3.2 3B LoRa & 0.622 & 0.630 & 0.621 & 0.623 & 1.000 \\
DeepSeek-R1-Distill-Qwen-1.5B LoRa &0.544 & 0.692 & 0.539 & 0.553 & 0.995 \\
Phi-4-mini-reasoning-3.8B LoRa& 0.635 & 0.635 & 0.635 & 0.635 & 0.988 \\ 
\midrule
\noalign{\vskip -3pt}
\multicolumn{5}{c}{\textbf{ISRR}} \\
\noalign{\vskip -3pt}
\midrule
gpt-4o-mini & 0.632 & 0.657 & 0.717 & 0.657 & 1.000\\
llama 3.2 3B & 0.529 & 0.530 & 0.530 & 0.529 & 0.988 \\ 
DeepSeek-R1-Distill-Qwen-1.5B &0.450 & 0.716 & 0.498 & 0.484 & 0.790 \\
Phi-4-mini-reasoning-3.8B & 0.600 & 0.490 & 0.638 & 0.605 & 0.988 \\
Qwen2.5-3B-Instruct &0.540 & 0.475 & 0.549 & 0.542 & 1.000 \\
Qwen2.5-3B-Instruct LoRa & 0.557 & 0.523 & 0.562 & 0.558 & 0.995\\ 
llama 3.2 3B LoRa & 0.589 & 0.505 & 0.612 & 0.593 & 1.000\\
DeepSeek-R1-Distill-Qwen-1.5B LoRa &0.491 & 0.731 & 0.512 & 0.515 & 0.980 \\
Phi-4-mini-reasoning-3.8B LoRa&0.579 & 0.360 & 0.706 & 0.604 & 0.998 \\ 
\bottomrule
\end{tabular}
\caption{Results on TREC test set. Macro F\textsubscript{1}, Precision and Recall.}
\label{tab:trec_metrics}
\end{table*}
\FloatBarrier

\maketitle
\begin{table*}[t]
\small
\centering
\begin{tabular}{lccccc}
\toprule
\textbf{Model} & \textbf{F\textsubscript{1}} & \textbf{Recall} & \textbf{Precision} & \textbf{Accuracy} & \textbf{\% Valid} \\
\midrule
\noalign{\vskip -3pt}
\multicolumn{5}{c}{\textbf{NLR}} \\
\noalign{\vskip -3pt}
\midrule
gpt-4o-mini & 0.799 & 0.807 & 0.797 & 0.797 & 1.000\\
llama 3.2 3B &0.427 & 0.432 & 0.435 & 0.432 & 1.000 \\ 
Qwen2.5-3B-Instruct & 0.498 & 0.495 & 0.520 & 0.496 & 0.978\\
DeepSeek-R1-Distill-Qwen-1.5B & 0.521 & 0.523 & 0.540 & 0.522 & 0.983 \\
Phi-4-mini-reasoning-3.8B & 0.690 & 0.691 & 0.693 & 0.691 & 0.995\\
llama 3.2 3B LoRa & 0.624 & 0.645 & 0.664 & 0.645 & 1.000 \\
Qwen2.5-3B-Instruct LoRa &0.652 & 0.658 & 0.653 & 0.657 & 1.000\\ 
DeepSeek-R1-Distill-Qwen-1.5B LoRa & 0.526 & 0.530 & 0.529 & 0.530 & 1.000\\
Phi-4-mini-reasoning-3.8B LoRa& 0.657 & 0.658 & 0.665 & 0.657 & 1.000\\ 
\midrule
\noalign{\vskip -3pt}
\multicolumn{5}{c}{\textbf{SSR}} \\
\noalign{\vskip -3pt}
\midrule
gpt-4o-mini & 0.799 & 0.800 & 0.800 & 0.800 & 1.000\\
llama 3.2 3B &0.413 & 0.458 & 0.517 & 0.461 & 0.960 \\ 
Qwen2.5-3B-Instruct & 0.630 & 0.626 & 0.653 & 0.626 & 0.995 \\
DeepSeek-R1-Distill-Qwen-1.5B & 0.420 & 0.423 & 0.430 & 0.424 & 0.790\\
Phi-4-mini-reasoning-3.8B &0.582 & 0.585 & 0.586 & 0.585 & 0.988 \\
llama 3.2 3B LoRa & 0.424 & 0.522 & 0.487 & 0.522 & 1.000\\
Qwen2.5-3B-Instruct LoRa &0.639 & 0.641 & 0.649 & 0.643 & 0.980 \\ 
DeepSeek-R1-Distill-Qwen-1.5B LoRa &0.447 & 0.469 & 0.463 & 0.471 & 0.988 \\
Phi-4-mini-reasoning-3.8B LoRa& 0.640 & 0.647 & 0.665 & 0.648 & 0.953\\ 
\midrule
\noalign{\vskip -3pt}
\multicolumn{5}{c}{\textbf{TAR}} \\
\noalign{\vskip -3pt}
\midrule
gpt-4o-mini & 0.758 & 0.759 & 0.762 & 0.762 & 1.000\\
llama 3.2 3B & 0.466 & 0.482 & 0.504 & 0.482 & 0.960\\ 
Qwen2.5-3B-Instruct & 0.563 & 0.567 & 0.611 & 0.567 & 0.983\\
DeepSeek-R1-Distill-Qwen-1.5B &0.373 & 0.401 & 0.396 & 0.402 & 0.870 \\
Phi-4-mini-reasoning-3.8B & 0.574 & 0.574 & 0.592 & 0.573 & 0.978\\
llama 3.2 3B LoRa & 0.465 & 0.555 & 0.561 & 0.555 & 1.000 \\
Qwen2.5-3B-Instruct LoRa & 0.541 & 0.575 & 0.578 & 0.575 & 1.000 \\ 
DeepSeek-R1-Distill-Qwen-1.5B LoRa &0.368 & 0.420 & 0.425 & 0.420 & 1.000 \\
Phi-4-mini-reasoning-3.8B LoRa&0.589 & 0.612 & 0.649 & 0.613 & 1.000 \\ 
\midrule
\noalign{\vskip -3pt}
\multicolumn{5}{c}{\textbf{ISRR}} \\
\noalign{\vskip -3pt}
\midrule
gpt-4o-mini & 0.789 & 0.792 & 0.790 & 0.790 & 1.000\\
llama 3.2 3B &0.379 & 0.410 & 0.506 & 0.411 & 0.998 \\ 
Qwen2.5-3B-Instruct & 0.587 & 0.594 & 0.634 & 0.596 & 0.965\\
DeepSeek-R1-Distill-Qwen-1.5B &0.360 & 0.384 & 0.384 & 0.383 & 0.685 \\
Phi-4-mini-reasoning-3.8B &0.577 & 0.575 & 0.594 & 0.575 & 0.988 \\
llama 3.2 3B LoRa & 0.439 & 0.519 & 0.512 & 0.521 & 0.993\\
Qwen2.5-3B-Instruct LoRa &0.579 & 0.585 & 0.601 & 0.587 & 0.945 \\ 
DeepSeek-R1-Distill-Qwen-1.5B LoRa &0.395 & 0.428 & 0.461 & 0.432 & 0.902 \\
Phi-4-mini-reasoning-3.8B LoRa& 0.596 & 0.618 & 0.627 & 0.618 & 0.995\\ 
\bottomrule
\end{tabular}
\caption{Results on MedNLI test set. Macro F\textsubscript{1}, Macro Precision, and Macro Recall}
\label{tab:med_metrics}
\end{table*}
\FloatBarrier

\begin{table*}[t]
\small
\centering
\begin{tabular}{lccccc}
\toprule
\textbf{Model} & \textbf{F\textsubscript{1}} & \textbf{Recall} & \textbf{Precision} & \textbf{Accuracy}& \textbf{\% Valid} \\
\midrule
\textbf{Direct}\\
\midrule
gpt-4o-mini & 0.643 & 0.696 & 0.630 & 0.644 & 1.000 \\
\midrule
llama 3.2 3B &0.490 & 0.817 & 0.521 & 0.532 & 0.962 \\ 
Qwen2.5-3B-Instruct & 0.501 & 0.327 & 0.533 & 0.519 & 0.990 \\
DeepSeek-R1-Distill-Qwen-1.5B &0.560 & 0.414 & 0.601 & 0.571 & 0.978 \\
Phi-4-mini-reasoning-3.8B &0.589 & 0.384 & 0.701 & 0.610 & 1.000 \\
\midrule
llama 3.2 3B LoRa & 0.490 & 0.696 & 0.506 & 0.508 & 1.000 \\
Qwen2.5-3B-Instruct LoRa & 0.521 & 0.424 & 0.537 & 0.525 & 0.944 \\ 
DeepSeek-R1-Distill-Qwen-1.5B LoRa &0.502 & 0.318 & 0.545 & 0.523 & 0.972 \\
Phi-4-mini-reasoning-3.8B LoRa&0.542 & 0.313 & 0.650 & 0.573 & 0.998\\ 
\midrule
\textbf{NLR} \\
\midrule
gpt-4o-mini & 0.800 & 0.815 & 0.776 & 0.800 & 1.000\\
\midrule
llama 3.2 3B &0.545 & 0.721 & 0.542 & 0.557 & 0.980 \\ 
Qwen2.5-3B-Instruct & 0.575 & 0.612 & 0.571 & 0.576 & 1.000 \\
DeepSeek-R1-Distill-Qwen-1.5B &0.562 & 0.431 & 0.599 & 0.570 & 0.954 \\
Phi-4-mini-reasoning-3.8B &0.680 & 0.645 & 0.693 & 0.680 & 0.994 \\
\midrule
llama 3.2 3B LoRa & 0.701 & 0.660 & 0.721 & 0.702 & 1.000 \\
Qwen2.5-3B-Instruct LoRa &0.631 & 0.540 & 0.665 & 0.634 & 1.000 \\ 
DeepSeek-R1-Distill-Qwen-1.5B LoRa &0.647 & 0.552 & 0.687 & 0.650 & 1.000 \\
Phi-4-mini-reasoning-3.8B LoRa&0.729 & 0.687 & 0.750 & 0.729 & 0.998 \\
\midrule
\textbf{SSR} \\
\midrule
gpt-4o-mini & 0.759 & 0.727 & 0.832 & 0.760 & 1.000\\
\midrule
llama 3.2 3B &0.441 & 0.819 & 0.492 & 0.495 & 0.808 \\ 
Qwen2.5-3B-Instruct & 0.568 & 0.708 & 0.556 & 0.575 & 0.884\\
DeepSeek-R1-Distill-Qwen-1.5B &0.479 & 0.660 & 0.494 & 0.493 & 0.754 \\
Phi-4-mini-reasoning-3.8B &0.598 & 0.776 & 0.578 & 0.608 & 0.776 \\
\midrule
llama 3.2 3B LoRa & 0.638 & 0.606 & 0.651 & 0.638 & 0.978\\
Qwen2.5-3B-Instruct LoRa &0.666 & 0.598 & 0.696 & 0.667 & 0.992 \\ 
DeepSeek-R1-Distill-Qwen-1.5B LoRa & 0.594 & 0.619 & 0.588 & 0.594 & 0.980 \\
Phi-4-mini-reasoning-3.8B LoRa&0.662 & 0.710 & 0.647 & 0.663 & 0.996\\ 
\midrule
\textbf{TAR} \\
\midrule
gpt-4o-mini & 0.798 & 0.794 & 0.804 & 0.798 & 1.000\\
\midrule
llama 3.2 3B &0.541 & 0.753 & 0.545 & 0.559 & 0.980 \\ 
Qwen2.5-3B-Instruct &0.580 & 0.468 & 0.613 & 0.586 & 1.000 \\
DeepSeek-R1-Distill-Qwen-1.5B &0.528 & 0.443 & 0.527 & 0.532 & 0.894 \\
Phi-4-mini-reasoning-3.8B & 0.686 & 0.669 & 0.689 & 0.686 & 0.976\\
\midrule
llama 3.2 3B LoRa & 0.682 & 0.684 & 0.681 & 0.682 & 1.000 \\
Qwen2.5-3B-Instruct LoRa & 0.667 & 0.663 & 0.668 & 0.667 & 0.998 \\ 
DeepSeek-R1-Distill-Qwen-1.5B LoRa &0.584 & 0.560 & 0.586 & 0.584 & 0.996 \\
Phi-4-mini-reasoning-3.8B LoRa&0.719 & 0.668 & 0.746 & 0.719 & 0.998 \\ 
\midrule
\textbf{ISRR} \\
\midrule
gpt-4o-mini & 0.780 & 0.890 & 0.648 & 0.784 & 1.000\\
\midrule
llama 3.2 3B & 0.475 & 0.806 & 0.510 & 0.518 & 0.996 \\ 
Qwen2.5-3B-Instruct & 0.511 & 0.688 & 0.518 & 0.524 & 1.000\\
DeepSeek-R1-Distill-Qwen-1.5B &0.508 & 0.744 & 0.520 & 0.530 & 0.706 \\
Phi-4-mini-reasoning-3.8B &0.672 & 0.690 & 0.665 & 0.672 & 0.994 \\
\midrule
llama 3.2 3B LoRa & 0.674 & 0.574 & 0.722 & 0.677 & 0.998\\
Qwen2.5-3B-Instruct LoRa & 0.641 & 0.466 & 0.744 & 0.653 & 0.998 \\
DeepSeek-R1-Distill-Qwen-1.5B LoRa &0.552 & 0.348 & 0.644 & 0.575 & 0.992 \\
Phi-4-mini-reasoning-3.8B LoRa&0.700 & 0.584 & 0.768 & 0.704 & 1.000 \\
\bottomrule
\end{tabular}
\caption{Results on the NLI4CT test set. Macro F\textsubscript{1}, Precision, and Recall are reported with entailment as the positive label.}
\label{tab:metrics}
\end{table*}
\FloatBarrier

\begin{table*}[h!]
\centering
\small
\setlength{\tabcolsep}{2pt}
\begin{tabular}{lcccccccccc}
\toprule
\textbf{Class} & \textbf{Base SSR} & \textbf{LoRa SSR} & \textbf{Base NLR} & \textbf{LoRa NLR} & \textbf{Base TAR} & \textbf{LoRa TAR} & \textbf{Base ISRR} & \textbf{LoRa ISRR} & \textbf{Base ZS} & \textbf{LoRa ZS} \\
\midrule
Quantitative Comp & 0.428 & 0.652 & 0.565 & 0.789 & 0.499 & 0.723 & 0.461 & 0.727 & 0.508 & 0.455 \\
QuantD       & 0.481 & 0.620 & 0.510 & 0.616 & 0.533 & 0.578 & 0.521 & 0.534 & 0.542 & 0.539 \\
Lexical Equivalence & 0.403 & 0.665 & 0.605 & 0.690 & 0.566 & 0.652 & 0.440 & 0.671 & 0.493 & 0.532 \\
Clinical      & 0.463 & 0.592 & 0.478 & 0.677 & 0.497 & 0.668 & 0.417 & 0.667 & 0.442 & 0.477 \\
World‐Knowledge  & 0.394 & 0.617 & 0.583 & 0.688 & 0.529 & 0.679 & 0.466 & 0.625 & 0.453 & 0.503 \\
Evidence      & 0.421 & 0.658 & 0.525 & 0.734 & 0.625 & 0.702 & 0.474 & 0.731 & 0.507 & 0.429 \\
\bottomrule
\end{tabular}
\caption{Combined F1 scores for each class across all experimental settings with Llama-3.2-3B-Instruct on the NLI4CT test set. ZS = Zero-Shot.}
\end{table*}

\begin{table*}[h!]
\centering
\small
\setlength{\tabcolsep}{2pt}
\begin{tabular}{lcccccccccc}
\toprule
\textbf{Class} & \textbf{Base SSR} & \textbf{LoRa SSR} & \textbf{Base NLR} & \textbf{LoRa NLR} & \textbf{Base TAR} & \textbf{LoRa TAR} & \textbf{Base ISRR} & \textbf{LoRa ISRR} & \textbf{Base ZS} & \textbf{LoRa ZS} \\
\midrule
Quantitative Comp & 0.639 & 0.681 & 0.594 & 0.694 & 0.650 & 0.716 & 0.520 & 0.702 & 0.501 & 0.582 \\
QuantD & 0.522 & 0.684 & 0.547 & 0.543 & 0.485 & 0.627 & 0.516 & 0.561 & 0.470 & 0.477 \\
Lexical Equivalence & 0.564 & 0.660 & 0.563 & 0.637 & 0.542 & 0.643 & 0.487 & 0.635 & 0.502 & 0.497 \\
Clinical & 0.508 & 0.648 & 0.588 & 0.631 & 0.618 & 0.643 & 0.493 & 0.626 & 0.527 & 0.510 \\
World‐Knowledge & 0.513 & 0.636 & 0.595 & 0.660 & 0.646 & 0.724 & 0.616 & 0.611 & 0.386 & 0.575 \\
Evidence & 0.483 & 0.631 & 0.537 & 0.637 & 0.569 & 0.744 & 0.501 & 0.689 & 0.520 & 0.534 \\
\bottomrule
\end{tabular}
\caption{Combined F1 scores for each class across all experimental settings with Qwen2.5-3B-Instruct on the NLI4CT test set. ZS = Zero-Shot.}
\end{table*}

\begin{table*}[h!]
\centering
\small
\setlength{\tabcolsep}{2pt}
\begin{tabular}{lcccccccccc}
\toprule
\textbf{Class} & \textbf{Base SSR} & \textbf{LoRa SSR} & \textbf{Base NLR} & \textbf{LoRa NLR} & \textbf{Base TAR} & \textbf{LoRa TAR} & \textbf{Base ISRR} & \textbf{LoRa ISRR} & \textbf{Base ZS} & \textbf{LoRa ZS} \\
\midrule
Quantitative Comp & 0.454 & 0.672 & 0.660 & 0.637 & 0.547 & 0.590 & 0.567 & 0.573 & 0.631 & 0.541 \\
QuantD & 0.486 & 0.570 & 0.483 & 0.608 & 0.483 & 0.578 & 0.450 & 0.526 & 0.533 & 0.390 \\
Lexical Equivalence & 0.454 & 0.524 & 0.527 & 0.627 & 0.531 & 0.596 & 0.541 & 0.523 & 0.557 & 0.448 \\
Clinical & 0.433 & 0.594 & 0.615 & 0.621 & 0.550 & 0.563 & 0.451 & 0.500 & 0.558 & 0.563 \\
World‐Knowledge & 0.613 & 0.566 & 0.463 & 0.578 & 0.470 & 0.482 & 0.533 & 0.521 & 0.565 & 0.537 \\
Evidence & 0.472 & 0.611 & 0.533 & 0.680 & 0.490 & 0.645 & 0.410 & 0.631 & 0.550 & 0.539 \\
\bottomrule
\end{tabular}
\caption{Combined F1 scores for each class across all experimental settings with DeepSeek-R1-Distill-Qwen-1.5B on the NLI4CT test set. ZS = Zero-Shot.}
\end{table*}

\begin{table*}[h!]
\centering
\small
\setlength{\tabcolsep}{2pt}
\begin{tabular}{lcccccccccc}
\toprule
\textbf{Class} & \textbf{Base SSR} & \textbf{LoRa SSR} & \textbf{Base NLR} & \textbf{LoRa NLR} & \textbf{Base TAR} & \textbf{LoRa TAR} & \textbf{Base ISRR} & \textbf{LoRa ISRR} & \textbf{Base ZS} & \textbf{LoRa ZS} \\
\midrule
Quantitative Comp & 0.600 & 0.668 & 0.696 & 0.736 & 0.712 & 0.775 & 0.685 & 0.749 & 0.598 & 0.591 \\
QuantD & 0.723 & 0.638 & 0.669 & 0.662 & 0.628 & 0.649 & 0.702 & 0.681 & 0.548 & 0.496 \\
Lexical Equivalence & 0.556 & 0.650 & 0.656 & 0.725 & 0.684 & 0.717 & 0.627 & 0.695 & 0.619 & 0.488 \\
Clinical & 0.579 & 0.626 & 0.694 & 0.695 & 0.662 & 0.661 & 0.668 & 0.677 & 0.608 & 0.634 \\
World‐Knowledge & 0.574 & 0.701 & 0.713 & 0.713 & 0.648 & 0.676 & 0.602 & 0.693 & 0.545 & 0.454 \\
Evidence & 0.579 & 0.636 & 0.651 & 0.772 & 0.739 & 0.759 & 0.677 & 0.737 & 0.593 & 0.554 \\
\bottomrule
\end{tabular}
\caption{Combined F1 scores for each class across all experimental settings with Phi-4-mini-reasoning on the NLI4CT test set. ZS = Zero-Shot.}
\end{table*}
\FloatBarrier

\begin{table}[H]
\centering
\small
\setlength{\tabcolsep}{2pt}
\begin{tabular}{lccccc}
\toprule
\textbf{Class} & \textbf{ISRR} & \textbf{TAR} & \textbf{NLR} & \textbf{SSR} & \textbf{ZS} \\
\midrule
Quantitative Comp & 0.841 & 0.824 & 0.824 & 0.797 & 0.634 \\
QuantD & 0.652 & 0.768 & 0.810 & 0.767 & 0.623 \\
Lexical Equivalence & 0.798 & 0.768 & 0.778 & 0.711 & 0.619 \\
Clinical & 0.751 & 0.781 & 0.788 & 0.771 & 0.634 \\
World‐Knowledge & 0.778 & 0.785 & 0.772 & 0.762 & 0.643 \\
Evidence & 0.805 & 0.868 & 0.809 & 0.727 & 0.618 \\
\bottomrule
\end{tabular}
\caption{Combined F1 scores for each class across all experimental settings with GPT-4o mini on the NLI4CT test set. ZS = Zero-Shot.}
\end{table}

\begin{table}[H]
\centering
\small
\setlength{\tabcolsep}{4pt}
\begin{tabular}{@{}lccc@{}}
\toprule
\textbf{Reasoning Type} & \textbf{Factor} & \textbf{Partial $\eta^2$} & \textbf{$p$-value} \\
\midrule
Clinical     & Prompt & 0.332 & 0.0118 \\
Clinical     & LoRA  & 0.289 & 0.0012 \\
Clinical     & Model & 0.489 &$ < $0.001 \\
World‐Knowledge & Prompt & 0.329 & 0.0125 \\
World‐Knowledge  & LoRA  & 0.186 & 0.0122 \\
World‐Knowledge  & Model & 0.314 & 0.0078 \\
Evidence   & Prompt & 0.435 & 0.0011 \\
Evidence    & LoRA  & 0.467 &$ < $0.001 \\
Evidence    & Model & 0.395 & 0.0012 \\
Quantitative Comp & Prompt & 0.333 & 0.0117 \\
Quantitative Comp & LoRA  & 0.293 & 0.0011 \\
Quantitative Comp & Model & 0.331 & 0.0054 \\
QuantD  & Prompt & 0.300 & 0.0223 \\
QuantD  & LoRA  & 0.067 & 0.1472 \\
QuantD  & Model & 0.526 &$ < $0.001 \\
Lexical Equivalence    & Prompt & 0.390 & 0.0033 \\
Lexical Equivalence    & LoRA  & 0.261 & 0.0024 \\
Lexical Equivalence    & Model & 0.395 & 0.0013 \\
\bottomrule
\end{tabular}
\caption{Partial $\eta^2$ and $p$-values from fixed-effects Type II ANOVA models fitted separately for each reasoning type, with prompt, LoRA status, and model identity as predictors. For each reasoning type (Clinical, Common Sense, Existence, Numerical Comparison), we fitted a separate ordinary least squares (OLS) regression model predicting macro-F$1$ from three categorical predictors: prompt strategy, LoRA adaptation status, and model identity. We conducted a Type II ANOVA on each model to isolate the marginal effect of each factor, and computed partial eta-squared ($\eta^2\text{partial}$) to estimate the proportion of residual-adjusted variance attributable to each. This decomposition quantifies the unique contribution of each variable to performance variation within individual reasoning subdomains.}
\label{tab:per-reasoning-anova}
\end{table}

\begin{table}[h!]
\small
\centering
\setlength{\tabcolsep}{2pt} 
\begin{tabular}{@{}p{3cm}ccp{1.5cm}@{}}
\toprule
\textbf{Class} & \textbf{Best Method} & \textbf{Best F\textsubscript{1}} & \textbf{Margin} \\
\midrule
Clinical & NLR & 0.625 & 0.0546 \\\midrule
World‐Knowledge & NLR & 0.624 & 0.0530 \\\midrule
Evidence & TAR & 0.668 & 0.0855 \\\midrule
Quantitative Comp & NLR & 0.671 & 0.0617 \\\midrule
QuantD & TAR & 0.593 & 0.0351 \\\midrule
Lexical Equivalence & NLR & 0.629 & 0.0583 \\
\bottomrule
\end{tabular}
\caption{The prompt with the highest F\textsubscript{1} score for each reasoning class, on the NLI4CT test set, and the margin over the average F\textsubscript{1} of other prompts.}
\label{tab:best-method-gap}
\end{table}

\begin{figure*}[h!tbp]
 \centering
 \includegraphics[width=\textwidth]{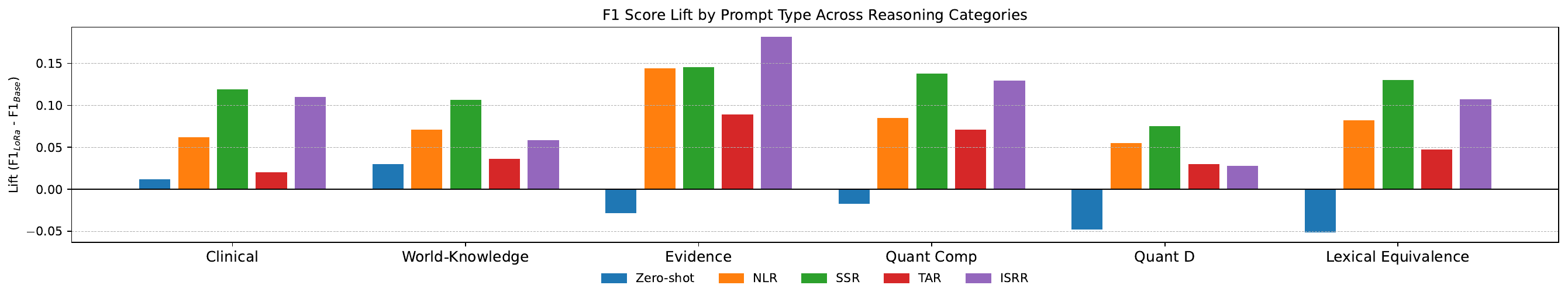}
 \caption{Impact of LoRA Tuning on F\textsubscript{1} Score by Prompt Strategy and Reasoning Type}
\label{fig:lift}
\end{figure*}

\begin{table}[H]
\centering
\small
\begin{tabular}{l c}
\toprule
\textbf{Prompt} & \textbf{Avg. Number of Demonstrations (k)} \\
\midrule
NLR  & 1.52 \\
SSR  & 1.46 \\
ISRR & 1.46 \\
TAR  & 1.55 \\
ZC  & 1.20 \\
\bottomrule
\end{tabular}
\caption{Number of demonstrations per prompt variant, in thousands.}
\label{tab:democount}
\end{table}

\begin{table}[H]
\centering
\small
\setlength{\tabcolsep}{3pt}
\begin{tabular}{@{}lccc@{}}
\toprule
\textbf{Factor} & \textbf{Eta$^2$} & \textbf{p-value} & \textbf{Effect Size} \\
\midrule
Prompt & 0.440 & 0.001 & Large \\
LoRA  & 0.357 & 0.0002 & Large \\
Model & 0.507 & 0.00006 & Very Large \\
\bottomrule
\end{tabular}
\caption{Overall effect of prompt, LoRA, and model on F\textsubscript{1} performance across all samples (averaged across reasoning types).
To assess the global contribution of prompt strategy, LoRA adaptation, and model architecture to performance variation we then fit an OLS model to F\textsubscript{1} score. with all predictors treated as fixed categorical effects. A Type II ANOVA was used to partition variance, and partial eta-squared values were calculated to quantify each factor’s unique explanatory power. This analysis estimates how much each design choice contributes to overall model performance across tasks.}
\label{tab:anova-global-effects}
\end{table}

\begin{table}[h!]
\centering
\small
\begin{tabular}{lc}
\toprule
\textbf{Reasoning Class} & \textbf{Mean F\textsubscript{1} (Lift vs. Base)} \\
\midrule
Clinical & 0.614 (0.065) \\
World‐Knowledge Inference & 0.612 (0.060) \\
Evidence & 0.653 (0.106) \\
Quantitative Comp & 0.663 (0.081) \\
QuantD & 0.579 (0.028) \\
Lexical Equivalence & 0.614 (0.063) \\
\bottomrule
\end{tabular}
\caption{Average F\textsubscript{1} scores of LoRa models by reasoning class, with lift compared to corresponding base models in parentheses.}
\label{tab:normal-vs-base-summary}
\end{table}

\begin{table*}[h]
\centering
\small
\begin{tabular}{llccc}
\toprule
\textbf{Method} & \textbf{Class} & \textbf{LoRa Mean F\textsubscript{1}} & \textbf{Base Mean F\textsubscript{1}} & \textbf{Lift} \\
\midrule
NLR & Clinical & 0.656 & 0.594 & 0.062 \\
NLR & World‐Knowledge Inference & 0.660 & 0.589 & 0.071 \\
NLR & Evidence & 0.706 & 0.562 & 0.144 \\
NLR & Quantitative Comp & 0.714 & 0.629 & 0.085 \\
NLR & QuantD & 0.607 & 0.552 & 0.055 \\
NLR & Lexical Equivalence & 0.670 & 0.588 & 0.082 \\
SSR & Clinical & 0.615 & 0.496 & 0.119 \\
SSR & World‐Knowledge Inference & 0.630 & 0.524 & 0.107 \\
SSR & Evidence & 0.634 & 0.489 & 0.145 \\
SSR & Quantitative Comp & 0.668 & 0.530 & 0.138 \\
SSR & QuantD & 0.628 & 0.553 & 0.075 \\
SSR & Lexical Equivalence & 0.625 & 0.494 & 0.131 \\
TAR & Clinical & 0.634 & 0.613 & 0.021 \\
TAR & World‐Knowledge Inference & 0.640 & 0.604 & 0.036 \\
TAR & Evidence & 0.713 & 0.623 & 0.089 \\
TAR & Quantitative Comp & 0.701 & 0.630 & 0.071 \\
TAR & QuantD & 0.608 & 0.578 & 0.030 \\
TAR & Lexical Equivalence & 0.652 & 0.604 & 0.048 \\
ISRR & Clinical & 0.618 & 0.507 & 0.110 \\
ISRR & World‐Knowledge Inference & 0.613 & 0.554 & 0.058 \\
ISRR & Evidence & 0.697 & 0.516 & 0.182 \\
ISRR & Quantitative Comp & 0.688 & 0.558 & 0.130 \\
ISRR & QuantD & 0.576 & 0.547 & 0.028 \\
ISRR & Lexical Equivalence & 0.631 & 0.524 & 0.107 \\
Zero-shot & Clinical & 0.546 & 0.534 & 0.012 \\
Zero-shot & World‐Knowledge Inference & 0.517 & 0.487 & 0.030 \\
Zero-shot & Evidence & 0.514 & 0.543 & -0.029 \\
Zero-shot & Quantitative Comp & 0.542 & 0.560 & -0.017 \\
Zero-shot & QuantD & 0.476 & 0.523 & -0.048 \\
Zero-shot & Lexical Equivalence & 0.491 & 0.543 & -0.052 \\
\bottomrule
\end{tabular}
\caption{Comparison of LoRa vs. base model performance (mean F\textsubscript{1}) across reasoning classes and prompt methods on the NLI4CT test set. The “Lift” column shows the performance gain of the LoRa-tuned model over its base counterpart.}
\label{tab:lift-analysis}
\end{table*}

\FloatBarrier
\subsection{Related Work Table}

\begin{table}[H]
\centering
\setlength{\tabcolsep}{2pt}
\scriptsize
\begin{tabular}{@{}p{2cm}p{3.7cm}p{2.3cm}@{}}
\toprule
\textbf{Study} & \textbf{Reasoning} & \textbf{Prompt strategies} \\
\midrule
\raggedright\cite{wen2025thinkpatterns} & Math, Commonsense & Monologue, Self-ask, Self-debate\\\midrule
\raggedright\cite{zhang2022automatic} &\raggedright Math, Commonsense, Symbolic & Manual vs.\ Auto CoT \\\midrule
\raggedright \cite{press2022measuring} & Multi-hop, Compositional & Self-ask vs.\ baseline \\\midrule
\raggedright \cite{mondorf2024beyond} & Math, Logical, Causal, Commonsense, Scientific, Social & CoT\\\midrule
\raggedright(Khalid et al. 2025) & Relational (spatial, temporal) & Zero-shot, CoT \\\midrule
\raggedright\cite{yu2025benchmarking} & Mathematical & CoT \\
\bottomrule
\end{tabular}
\caption{Cross-domain studies contrasting prompting strategies and reasoning types.}
\label{tab:cross_reasoning_studies}
\end{table}

\end{document}